\documentclass[10pt,twocolumn,letterpaper]{article}
\usepackage[dvipsnames,table]{xcolor}

\usepackage[most]{tcolorbox}
\usepackage[pagenumbers]{cvpr} 

\usepackage{array}
\usepackage{url}
\usepackage{graphicx}
\usepackage{amsmath}
\usepackage{booktabs}       
\usepackage{multirow}
\usepackage{nicematrix}
\usepackage{pifont}

\usepackage{algorithm} 
\usepackage{algpseudocode} 

\definecolor{colorYes}{RGB}{51,160,44}
\definecolor{colorNo}{RGB}{228,26,28} %
\newcommand{\cmark}{\textcolor{colorYes}{\ding{51}}}%
\newcommand{\xmark}{\textcolor{colorNo}{\ding{55}}}%

\usepackage[textwidth=2.5cm]{todonotes}

\definecolor{cvprblue}{rgb}{0.21,0.49,0.74}
\usepackage[pagebackref,breaklinks,colorlinks,allcolors=cvprblue]{hyperref}
\usepackage[capitalize]{cleveref}

\def\paperID{*****} 
\def\confName{CVPR}
\def\confYear{2026}

\newcommand{\ours}{\textsc{VSAS-Bench}}
\newcommand{\present}{{Present}}
\newcommand{\cumulative}{{Cumulative}}
\newcommand{\future}{{Future}}
\newcommand{\SWaccess}{\textsc{SW}}
\newcommand{\Uaccess}{\textsc{U}}
\newcommand{\SWUaccess}{\textsc{SW+U}}

\title{\ours{}: Real-Time Evaluation of Visual Streaming Assistant Models}

\author{Pavan Kumar Anasosalu Vasu\thanks{Equal contribution. \tt\scriptsize \{panasosaluvasu,cem\_koc,fartash\}@apple.com}, \
Cem Koc$^*$, 
Fartash Faghri$^*$, 
Chun-Liang Li,\\
Bo Feng, \
Zhengfeng Lai, \
Meng Cao, \
Oncel Tuzel, \
Hadi Pouransari$^*$
\\
Apple
}

\begin{document}
\maketitle
\begin{abstract}
Streaming vision-language models (VLMs) continuously generate responses given an instruction prompt and an online stream of input frames. This is a core mechanism for real-time visual assistants. Existing VLM frameworks predominantly assess models in offline settings. In contrast, the performance of a streaming VLM depends on additional metrics beyond pure video understanding, including proactiveness, which reflects the timeliness of the model’s responses, and consistency, which captures the robustness of its responses over time. To address this limitation, we propose \ours{}, a new framework and benchmark for Visual Streaming Assistants. In contrast to prior benchmarks that primarily employ single-turn question answering on video inputs, \ours{} features temporally dense annotations with over 18,000 annotations across diverse input domains and task types. We introduce standardized synchronous and asynchronous evaluation protocols, along with metrics that isolate and measure distinct capabilities of streaming VLMs. Using this framework, we conduct large-scale evaluations of recent video and streaming VLMs, analyzing the accuracy–latency trade-off under key design factors such as memory buffer length, memory access policy, and input resolution, yielding several practical insights. Finally, we show empirically that conventional VLMs can be adapted to streaming settings without additional training, and demonstrate that these adapted models outperform recent streaming VLMs. For example, Qwen3-VL-4B surpasses Dispider, the best streaming VLM on our benchmark by 3\% under asynchronous protocol. The benchmark and code will be available at \url{https://github.com/apple/ml-vsas-bench}.

\end{abstract}

\begin{figure}[ht]
\centering
\includegraphics[width=0.85\columnwidth]{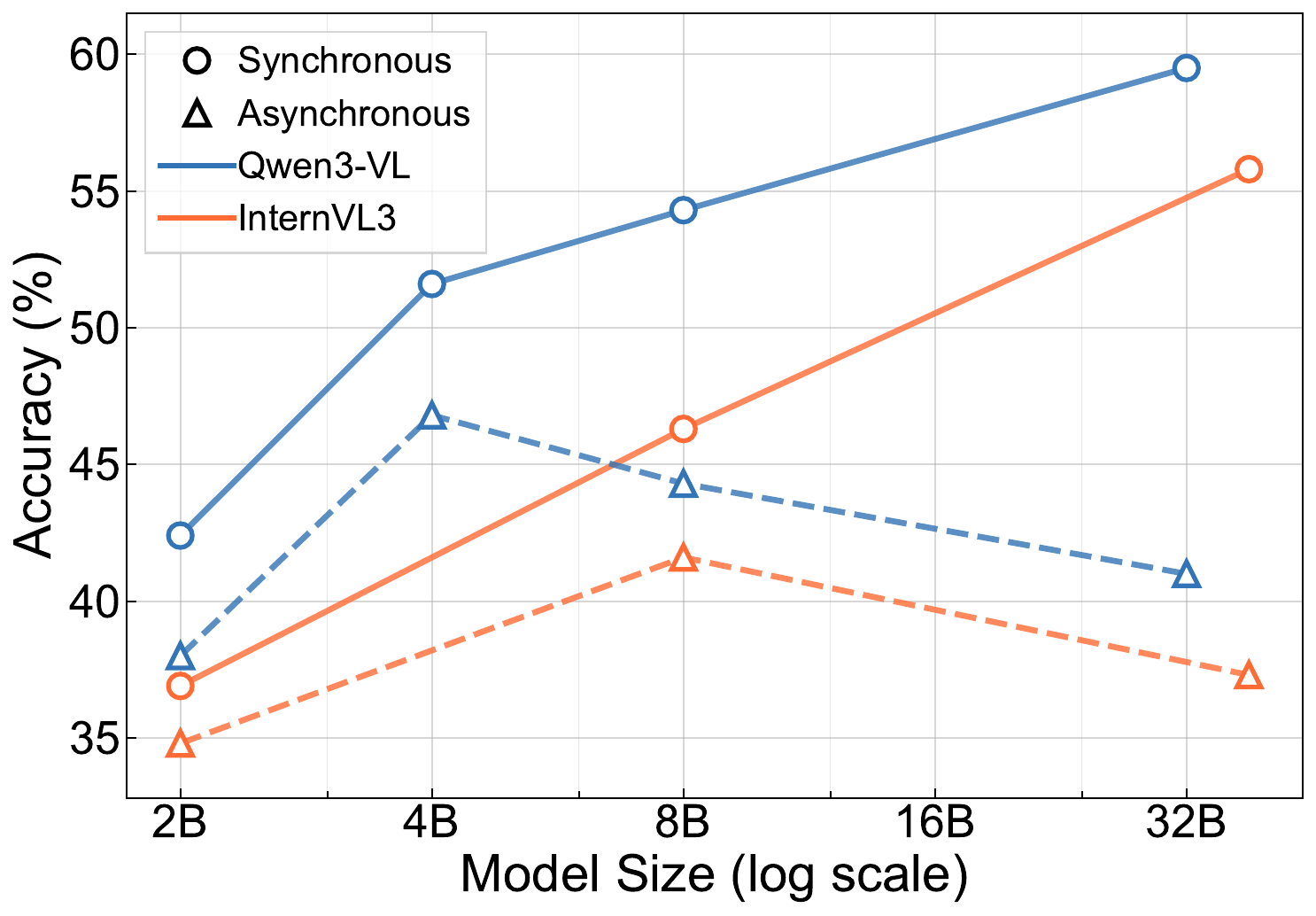}
\vspace*{-5pt}
\caption{\textbf{\ours{} accuracy uncovers advantage of small models as visual streaming assistants.} Under the synchronous evaluation protocol, models run in lockstep with the camera and larger models benefit from unlimited processing time. However, in the more realistic asynchronous evaluation, smaller models outperform because their high inference speed allows them to respond more rapidly and effectively to the live stream.}
\label{fig:qwen3_sync_async_gap}
\end{figure}

\section{Introduction}\label{sec:intro}

\begin{table}[ht]
\begingroup
\setlength{\tabcolsep}{1pt} 
    \centering
    \caption{Comparison to prior video streaming benchmarks. }
    \vspace*{-3pt}
    \label{tab:benchmark_comparison}
    \resizebox{1.0\linewidth}{!}{
    \begin{tabular}{l
    >{\centering\arraybackslash}p{2.2cm}
    >{\centering\arraybackslash}p{1.9cm}
    >{\centering\arraybackslash}p{1.9cm}
    >{\centering\arraybackslash}p{1.9cm}
    >{\centering\arraybackslash}p{1.9cm}}
        \toprule[1.2pt]
        
        \multirow{2}{*}{\textbf{Benchmark}}
        & \textbf{Long Horizon}
        & \textbf{Dense}
        & \textbf{Free-form}
        & \textbf{Async. Eval.}
        & \textbf{Time-aware}
        \\

        & \textbf{Reasoning}
        & \textbf{Annotations}
        & \textbf{Response}
        & \textbf{Protocol}
        & \textbf{Metrics}
        \\
        
        \midrule[1.2pt]
        
        Ego4D~\citep{grauman2022ego4d}   & \cmark &  \cmark & \xmark & \xmark &  \xmark\\
        COIN~\citep{tang2019coin}    & \xmark &  \cmark & \xmark & \xmark &  \xmark\\
        VideoLLM-Online~\citep{chen2024videollm}   & \xmark &  \cmark & \cmark & \xmark & \cmark\\
        VStream-QA~\citep{zhang2024flash}  & \xmark &  \xmark & \cmark & \xmark & \xmark \\
        E.T. Bench~\citep{liu2024etbench} & \cmark & \xmark & \cmark & \xmark & \xmark \\
        StreamingBench~\citep{lin2024streamingbench}  & \cmark &  \xmark & \xmark & \xmark & \xmark \\
        OVO-Bench~\citep{niu2025ovo}   & \cmark &  \xmark & \xmark & \xmark & \xmark \\
        \textbf{\ours{} (ours)}   & \cmark &  \cmark & \cmark & \cmark & \cmark\\
        \bottomrule[1.2pt]
    \end{tabular}
    }
    \vspace{-5pt}
\endgroup
\end{table}

Recent advances in large Vision-Language Models (VLMs) have greatly improved multimodal reasoning and visual understanding~\citep{qwen3vl,comanici2025gemini,hurst2024gpt}. However, streaming settings pose additional challenges: models must process continuous visual input with low latency and respond selectively to avoid cognitive overload. In applications such as hazard detection for visually impaired users, cooking assistance, and interface navigation, a streaming VLM must track evolving visual context, infer user intent, and provide timely, relevant feedback. More broadly, effective streaming VLMs must balance responsiveness and interpretability, delivering not only prompt responses but also contextual and actionable guidance.

In this paper, we propose a \textit{Visual Streaming Assistant (VSAS)} 
benchmark and address several gaps in the literature (see 
\cref{tab:benchmark_comparison}). Our \ours{} dataset consists of 92 videos of varying duration, designed to introduce tasks that require both short- and long-horizon temporal reasoning (see \cref{sec:dataset} for details). Notably, our benchmark features temporally dense annotations and evaluates responses for completeness. In contrast, common streaming benchmarks 
often reduce complex questions to simplified multi-choice or Yes/No formats. For 
example, OVO-Bench\citep{niu2025ovo} converts questions such as \textit{``What 
is their purpose of doing so?''} into a binary decision prompt. Such evaluations fail to measure the diversity and 
complexity of responses required for effective real-world visual assistance.

Beyond providing a benchmark with rich and dense annotations, we propose new evaluation protocols tailored to streaming settings. Existing benchmarks such as OVO-Bench~\citep{niu2025ovo}, StreamingBench~\citep{lin2024streamingbench}, and {E.T. Bench}~\citep{liu2024etbench} employ synchronous evaluation, wherein models operate in lockstep with the camera stream, allowing arbitrary processing latency between frames. Such time-agnostic setups fail to penalize models that are accurate yet computationally slow. To address this limitation, we introduce an \textit{asynchronous evaluation protocol} and time-aware accuracy metrics. In this setup, frames arrive at fixed intervals and stored in a bounded camera buffer. Inference is initiated on available frames whenever the model becomes free; if inference is delayed, older frames may be dropped, altering the effective temporal context.
Asynchronous protocol explicitly captures the trade-off between computational efficiency and temporal fidelity, an aspect overlooked by prior benchmarks. 
To the best of our knowledge, our benchmark is the first to enable accuracy measurements that capture the benefits of runtime design choices (\cref{sec:async_ablations}) and model-specific optimization techniques (\cref{sec:other_optimizations}). Details of the evaluation protocol and associated metrics are provided in \cref{sec:evaluation_protocol,sec:metrics}, respectively.

We benchmark several state-of-the-art video and streaming VLMs under both synchronous and asynchronous evaluation protocols, revealing clear trade-offs between processing speed and reasoning depth. Furthermore, we introduce a lightweight method for converting video VLMs into streaming models via a memory buffer and access policy that retrieve and assemble relevant context for inference (\Cref{sec:video_to_streaming}). Under the asynchronous protocol, existing streaming VLMs often underperform due to fine-tuning on narrow, task-specific domains, whereas our streaming-adapted video VLMs are training-free extensions of base models and hence retain broader generalization capabilities. The following is a summary of our contributions:

\begin{itemize}
\item We introduce \ours{} with temporally dense, temporally grounded, free-form annotations across diverse videos and tasks, and provide a unified evaluation of recent VLMs and streaming models.

\item We propose streaming-focused metrics, accuracy and latency, and an asynchronous evaluation protocol whose time-aware accuracy captures the trade-off between computational delay and temporal fidelity.

\item We show empirically that training-free adaptation offers a strong recipe for enabling streaming behavior in video VLMs, with adapted models often generalizing better than recent task-specific streaming VLMs.

\end{itemize}

\begin{figure*}[ht]
    \centering
    
    \begin{subfigure}[b]{0.35\textwidth}
        \includegraphics[width=\textwidth]{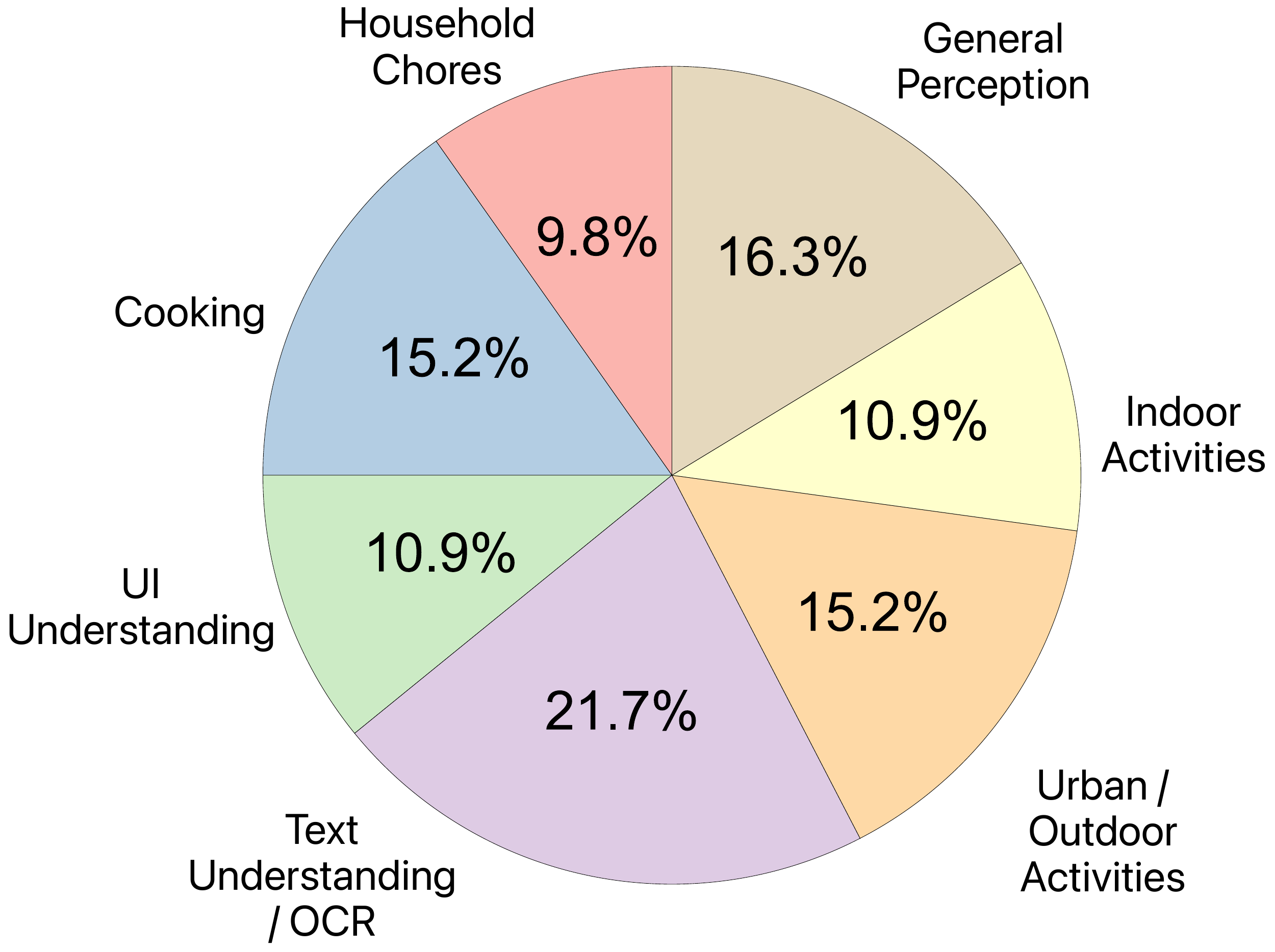}
        \caption{}
        \label{fig:data_distribution}
    \end{subfigure}
    \begin{subfigure}[b]{0.31\textwidth}
        \includegraphics[width=\textwidth]{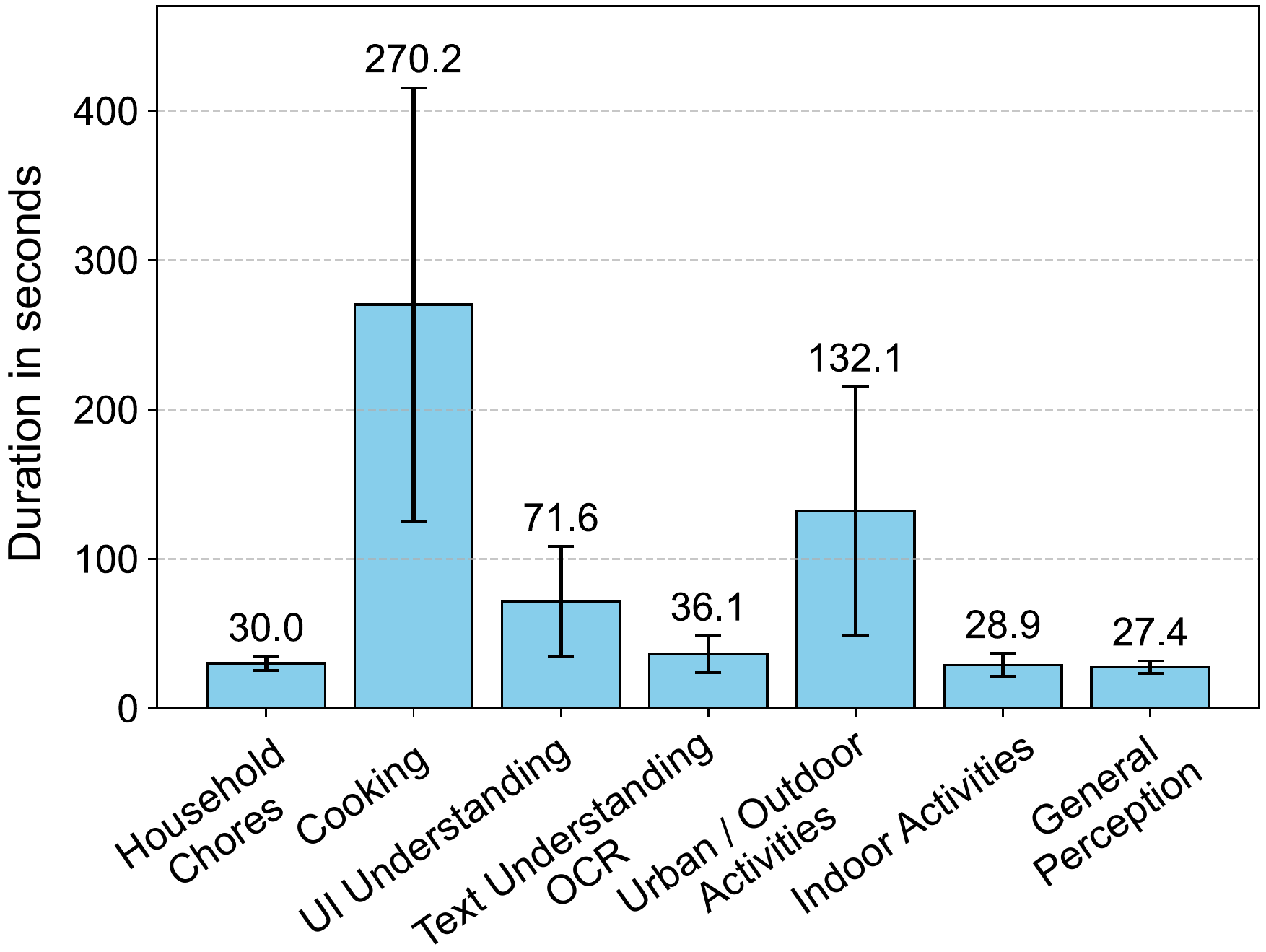}
        \caption{}
        \label{fig:video_duration_distribution}
    \end{subfigure}
    \begin{subfigure}[b]{0.31\textwidth}
        \includegraphics[width=\textwidth]{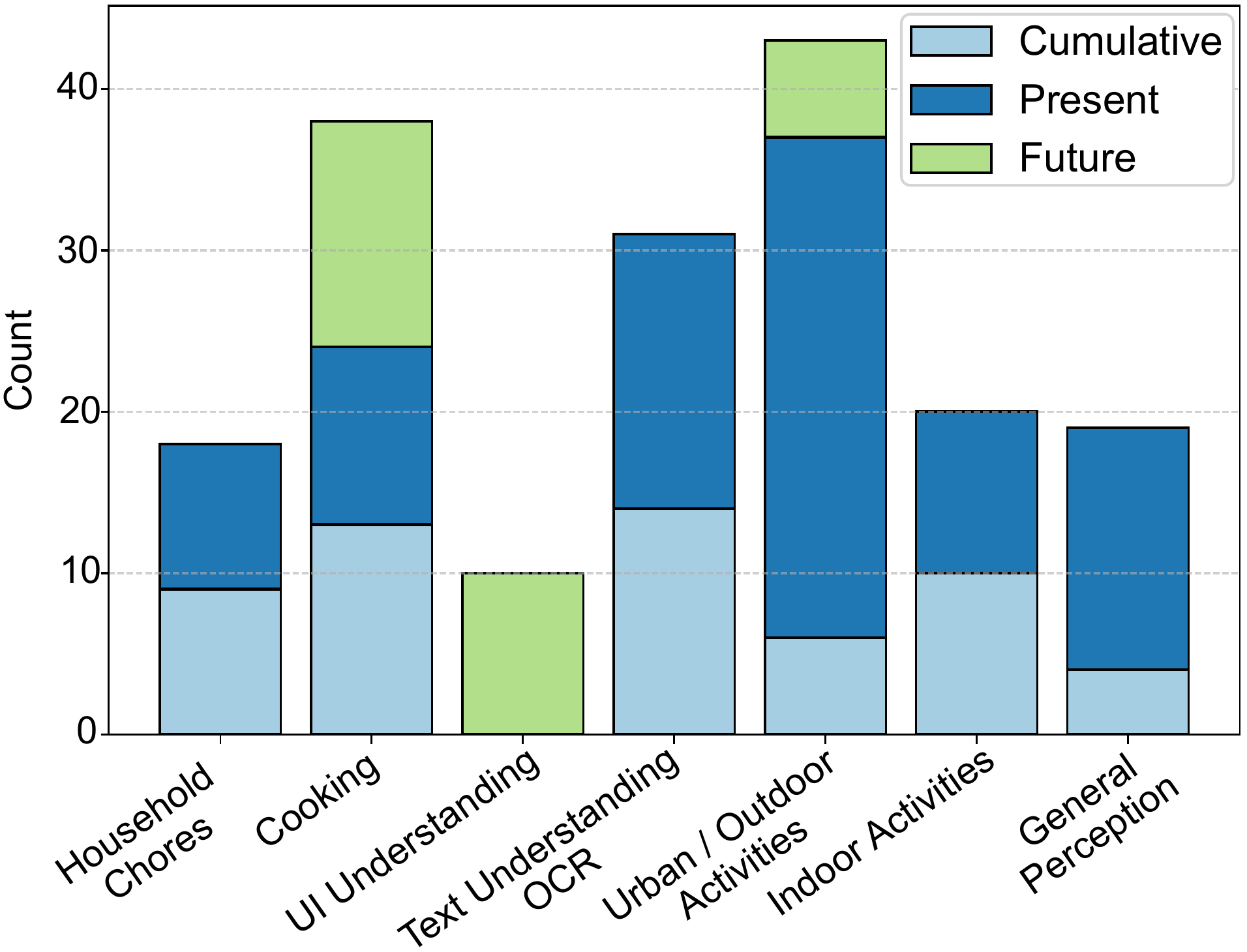}
        \caption{}
        \label{fig:task_distribution}
    \end{subfigure}
    \vspace*{-5pt}
    \caption{\textbf{\ours{} comprises densely annotated videos with a wide range of actions, varying durations, and reasoning horizons.}
    (a) Distribution of video categories, showing the diversity of tasks covered.
    (b)  Mean video duration (with standard deviation bars) for each video category.
    (c) Distribution of task types, highlighting the split between short- and long-horizon temporal reasoning.}
    \label{fig:benchmark_details}
    \vspace{-2mm}
\end{figure*}

\section{Related Work}\label{sec:related_works}

Most existing benchmarks in the literature focus on video understanding rather than streaming comprehension, and their annotations are typically sparse and often limited to a single question–answer pair per video or subclip~\citep{caba2015activitynet,xu2016msrvtt,cai2024temporalbenchbenchmarkingfinegrainedtemporal,xiao2021nextqa}.
{E.T. Bench}~\citep{liu2024etbench} evaluates models at a fine-grained event-level granularity, and other efforts like Video-MME~\citep{videomme}, LongVideoBench~\citep{wu2024longvideobench}, and EgoSchema~\citep{mangalam2023egoschema} extend to long-form videos. Furthermore,
Ego4D~\citep{grauman2022ego4d} and COIN~\citep{tang2019coin} contain hours of activity videos with temporally aligned dense annotations.
However, these benchmarks are not designed for streaming VLMs with user-assistant 
dialogue and interaction, where a single query may elicit multiple temporally grounded responses as the video unfolds. For example, even though Ego4D and COIN contain dense annotations, their tasks and metrics are not designed to evaluate free-form responses with time-aware metrics.

More recently, streaming-focused benchmarks have been introduced. 
VideoLLM-Online~\citep{chen2024videollm} repurposed instructional videos from COIN~\citep{tang2019coin} and egocentric recordings from Ego4D~\citep{grauman2022ego4d}, transforming their narrative annotations into a streaming, dialogue-based format. VideoLLM-Online is among the few streaming evaluation datasets featuring free-form natural language annotations with dense temporal granularity. 
VStream-QA~\citep{zhang2024flash} introduced timestamped questions for videos up to 60 minutes.
StreamingBench~\citep{lin2024streamingbench} consists of real-time visual 
understanding, omni-source understanding, and contextual understanding 
evaluations. Each video comes with five questions per video asked at varying 
timestamps.  OVO-Bench~\citep{niu2025ovo} further focused on various backward 
tracing, real-time understanding, and forward active responding tasks that 
emphasize the importance of video timestamps.  However, evaluations in 
OVO-Bench are based on accuracy metrics for multi-choice questions that do not 
require answering questions with long responses.

A wide range of open-source Vision-Language Models (VLMs) support video inference~\citep{qwen3vl, Qwen2.5-VL, Qwen2-VL, internvl3, llava-onevision, hong2024cogvlm2, marafioti2025smolvlm, fastvlm2025, yao2024minicpm, liu2024nvila}. However, these models are primarily designed for static, offline settings, where they typically process entire videos by uniformly sampling a fixed number of frames. While suitable for conventional video understanding benchmarks, this paradigm is incompatible with streaming scenarios, where frames arrive sequentially and must be processed in real time. The key distinction between video and streaming VLMs lies in how they manage temporal context and selective responding. Recent streaming models like FlashVStream~\citep{zhang2024flash}, VideoLLM-Online~\citep{chen2024videollm}, Dispider~\citep{qian2025dispider}, StreamBridge~\citep{wang2025streambridge}, and LiveCC~\citep{livecc} address this by introducing specialized memory mechanisms for handling continuous visual streams. However, such designs often rely on task-specific fine-tuning or retraining, limiting their generalization.
LiveCC is designed specifically for captioning and is therefore excluded from our evaluations. In contrast, we propose a simple, training-free approach to adapt video VLMs for streaming use via a lightweight memory buffer and access policy (\Cref{sec:video_to_streaming}), enabling them to preserve the broad generalization capabilities of their base models.

\begin{figure*}[ht]
    \centering
    \includegraphics[width=0.82\textwidth, height=0.75\textheight, keepaspectratio=true]{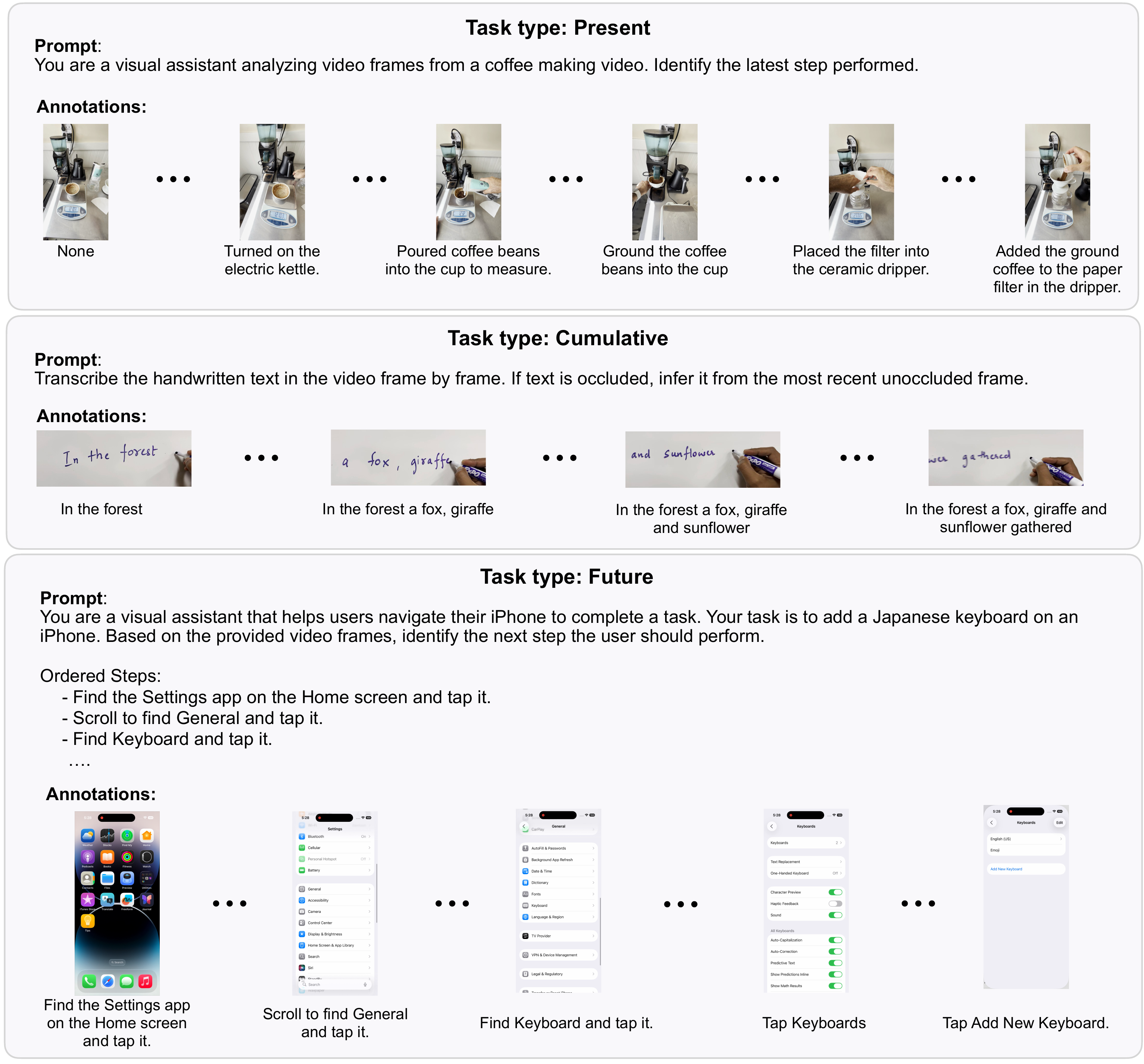}
    \vspace*{-5pt}
    \caption{\textbf{Examples of \ours{} task types with frame-level annotations.} \ours{} involves three task types, \present{} tasks, which focus on currently occurring events; \cumulative{} tasks, which require the model to reason over past events; and \future{} tasks, which focus on predicting upcoming events based on ongoing visual cues.}
    \label{fig:benchmark_annotations}
    \vspace{-2mm}
\end{figure*}

\begin{figure*}[ht]
    \centering
    \includegraphics[width=0.95\textwidth]{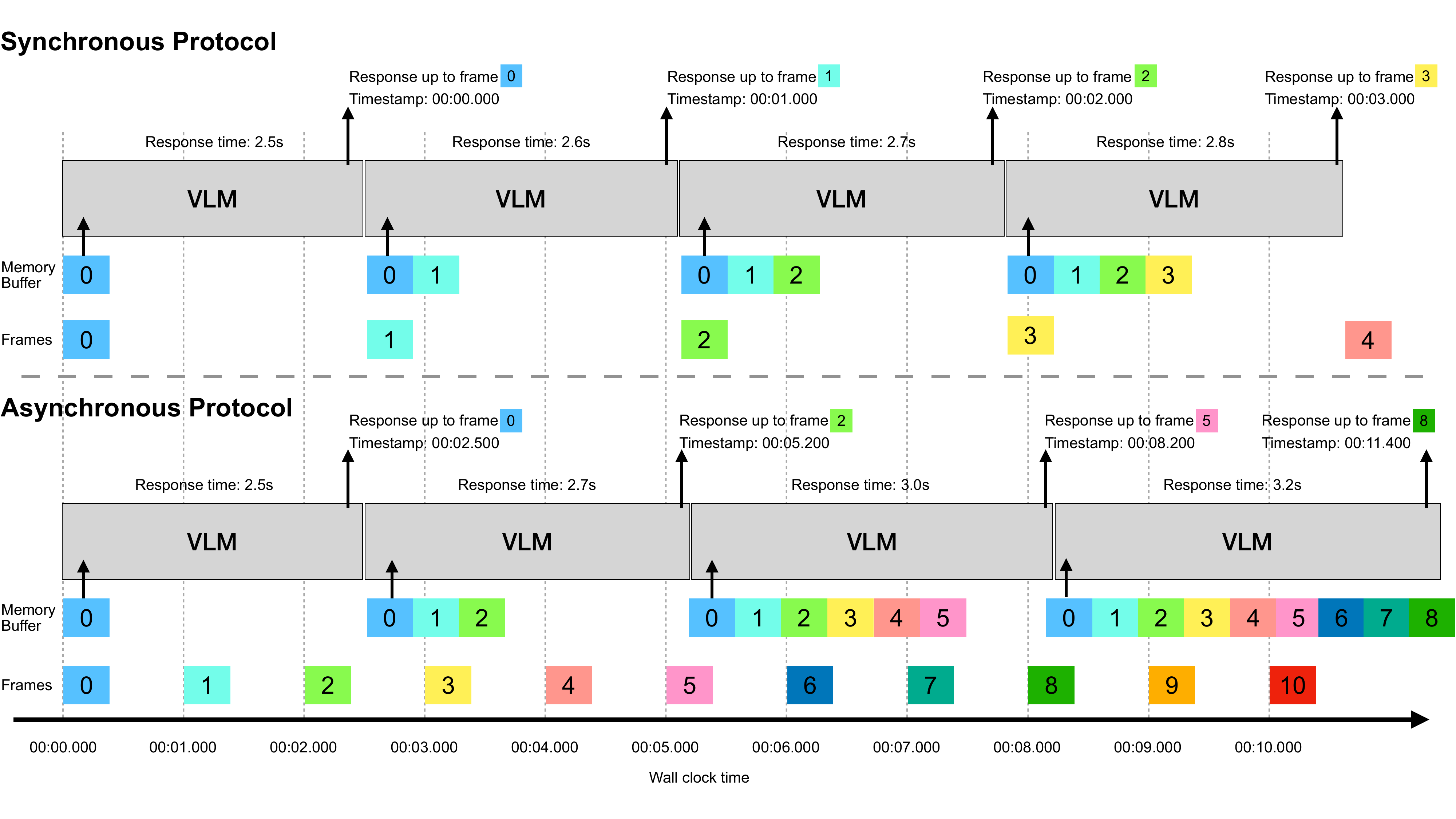}
    \vspace*{-10pt}
    \caption{\textbf{Traditional versus realistic evaluation protocols.} (top) 
    \textbf{Synchronous Protocol}: Frame generation is temporally aligned with 
    the VLM’s processing rate, ensuring camera input and model inference occur 
    in lockstep. (bottom) \textbf{Asynchronous Protocol}: Frame acquisition is 
    decoupled from inference; the VLM retrieves frames from a buffered queue, 
    emulating real-world streaming on edge devices.
}
    \label{fig:sync_async}
\end{figure*}

\section{Benchmark}

\ours{} targets proactive response settings, where the model receives a single initial prompt and can produce multiple responses as new visual input arrives. The next section details the benchmark design, including the task types and the two main evaluation protocols.

\subsection{Dataset and Annotations}\label{sec:dataset}

Our benchmark comprises 92 videos of varying durations, including 48 newly recorded videos that capture realistic use cases typical of streaming vision-language applications such as cooking activities, urban human actions, and UI interaction tasks.
The remaining 44 videos are sourced from existing benchmarks, namely STAR~\citep{wu2021star_situated_reasoning}, PerceptionTest~\citep{patraucean2023perceptiontest}, and YouCook2~\citep{youcook2}. 
\Cref{fig:data_distribution,fig:video_duration_distribution} show the distribution of video categories and durations.

The videos are annotated for three primary task types: \present{}, \cumulative{}, and \future{}. \Cref{fig:task_distribution} shows the distribution of these tasks across video categories. In the \present{} task, the queries pertain to events currently occurring in the stream. The \cumulative{} task involves prompts requiring the model to recollect and reason over past events observed so far. Finally, the \future{} task focuses on predicting or inferring upcoming events based on the ongoing visual context.

Each video is annotated at a temporal resolution of 1 frame per second (FPS). Initial annotations are generated with GPT-5~\citep{openai2025gpt5} (medium reasoning) and then reviewed by human experts to ensure correctness and consistency. In total, all 18k annotations underwent expert verification, with approximately 10\% requiring correction due to semantic inconsistencies. We further assessed annotation quality on a subset of $N=250$ samples spanning all three tasks by measuring GPT--human semantic agreement using cosine similarity of \texttt{all-mpnet-base-v2} embeddings, with thresholds calibrated on a control set. Human annotation reliability is high, with 96.7\% inter-annotator agreement. Additional implementation details and example prompts are provided in the Appendix. Unlike prior benchmarks such as OVO-Bench and StreamingBench, which formulate annotations as discrete queries, our benchmark uses free-form responses and explicitly specifies task details in the prompt to reduce ambiguity and prevent inference of unstated goals. For example, the prompt for the \future{} task in \cref{fig:benchmark_annotations}(bottom) specifies the sequence of actions required to complete the activity. This formulation ensures clarity in task intent and aligns with emerging agentic workflows where a larger server-side model generates high-level plans while smaller on-device models interpret visual context and respond accordingly.

\subsection{Evaluation Protocol}\label{sec:evaluation_protocol}

\subsubsection{Synchronous Protocol}
Most current video and streaming benchmarks adopt a synchronous evaluation protocol, where the model processes entire video segments or non-contiguous sub-segments within a stream. In this setting, there is no notion of a continuous camera or image stream, and inference latency does not influence the reported metrics. In prior benchmarks like OVO-Bench, even streaming VLMs are typically evaluated under this protocol, with frame generation temporally aligned to the model’s processing rate. As a result, slower models incur no penalty, since frames are streamed in lockstep with VLM model inference. While this setup is useful for assessing video understanding in an offline setting, it fails to reflect the dynamics of a true streaming environment. An illustration of this protocol appears in \cref{fig:sync_async}(top).

\subsubsection{Asynchronous Protocol}\label{sec:async_protocol}
In the asynchronous evaluation protocol, two primary processes are maintained: one simulates a continuous camera stream and the other hosts the VLM which consumes frames and generates responses asynchronously similar to a producer-consumer pattern. This setup reflects a realistic online scenario where the camera typically operates at a higher frame rate than the model’s inference speed. To bridge this rate mismatch, a \emph{camera buffer} stores the most recent frames and provides them to the VLM once it is ready for processing. Similar buffering mechanisms are commonly implemented in commercial camera systems to record short bursts of frames. We expose the camera buffer length as a configurable parameter, allowing researchers to emulate different deployment setups. As illustrated in \cref{fig:sync_async}(bottom), frames are streamed at 1 FPS. 
During periods when the VLM is busy, e.g., seconds 1–2, the camera buffer temporarily stores incoming frames and releases them once the model resumes processing at 2.5 seconds. 

Streaming VLMs are inherently compatible with and operate under asynchronous protocol, as their internal memory mechanisms maintain temporal continuity across streamed inputs. In contrast, conventional video models assume access to pre-segmented clips and thus cannot operate directly in real-time streaming settings. We introduce two external mechanisms: a model memory buffer and a memory policy which are detailed in \cref{sec:video_to_streaming} to adapt and run them as streaming-capable VLMs.

The asynchronous evaluation protocol is summarized in \cref{alg:async_protocol}. The camera buffer is implemented as a queue, ensuring decoupled operation between the camera and the VLM processes. To preserve the target frame rate during simulation, we compensate for timing variations introduced by video reading and buffer insertion operations, thereby maintaining a consistent streaming cadence. 

\subsection{Metrics}\label{sec:metrics}

\begin{algorithm}[t]
\caption{Asynchronous Protocol}
\label{alg:async_protocol}
\begin{algorithmic}[1]
\Require Video $V$, camera frame rate $f_c$, model inference time $t_m$, camera buffer size $B$
\State Initialize camera buffer $\mathcal{B} \leftarrow \emptyset$
\While{not isLastVideoFrame}
    \Comment{Camera Process}
    \State Read video frame $F_t$ at interval $1 / f_c$
    \State Append $F_t$ to $\mathcal{B}$ and discard oldest if $|\mathcal{B}| > B$
\EndWhile
\While{VLM process is active}
    \Comment{Model Process}
    \If{$\mathcal{B}$ is not empty}
        \State Retrieve most recent frame(s) $F_{t'}$ from $\mathcal{B}$
        \State Update model memory with $F_{t'}$
        \State Generate response $R_{t'} \leftarrow \text{VLM}(F_{t'}, \text{memory})$
    \Else
        \State Wait until next frame is available
    \EndIf
\EndWhile
\end{algorithmic}\label{alg:async_protocol}
\end{algorithm}

Below, we elaborate on the definitions of the metrics introduced for streaming VLMs in the \ours{}. We assume there are $K$ tasks in the benchmark. For a task $t$ with a video of $N_t$ seconds, we denote the model output at timestep $i$ as $R_i$ and the ground-truth caption as $G_i$ (for simplicity, we drop the task index $t$ in $R_i$ and $G_i$).

\noindent \textbf{Model response extrapolation.}
Streaming VLMs differ from conventional video models in that they may intentionally pause if they believe their previous response is still valid, or if they need to accumulate additional temporal context before responding. As a result, some timesteps may lack model responses. A similar situation can occur even for standard VLMs when evaluated under the asynchronous protocol: responses for some timesteps may be missing because of model's high latency in generation. To ensure a fair accuracy measure, we \emph{extrapolate} the model responses before computing the metrics. Specifically, for each timestep $i$ without a model response $R_i$ (or with a pause response by a streaming model), we use an available $R_j$ as model's response for the largest $j < i$. If no such $R_j$ exists, we consider an empty string as the response. After extrapolation, we assume $R_i$ is available at all timesteps.

\noindent \textbf{Mean Average Accuracy ($\mathcal{A}$).} 
We evaluate model correctness by comparing each generated response to the corresponding ground-truth annotation at every timestep. The frame-level accuracy measures both whether the model response is correct and whether it is generated at the right time. For each timestep $i$, we use an LLM judge to measure syntactic and semantic similarity between $R_i$ and $G_i$, denoted by \texttt{Judge}$(G_i, R_i) \in [0,1]$. These frame-level scores are averaged over all $N_t$ timesteps of a task video to obtain a per-task accuracy, as shown in \cref{eq:acc_per_video}. The overall benchmark accuracy metric, called \emph{mean average accuracy}, is computed as the mean of the $K$ per-task accuracies.

\vspace{-5mm}
\begin{gather}
    \mathcal{A}_t = \frac{1}{N_t} \sum_{i=1}^{N_t} \text{\texttt{Judge}}\left(G_{i}, R_{i}\right)
    \label{eq:acc_per_video} \tag{1} \\ 
    \mathcal{A} = \frac{1}{K} \sum_{t=1}^K \mathcal{A}_t
    \label{eq:overall_acc}\,, \tag{2}
\end{gather}
where $R_{i}$ is the model output for timestep $i$ of task $t$ and $G_{i}$ is the ground-truth caption for timestep $i$ of task $t$.

\noindent
\textbf{Mean Average Consistency ($\mathcal{C}$).} 
As visual assistants, streaming VLMs must not only be accurate but temporally consistent. A model may answer correctly at each timestep yet vary its responses enough to hinder usability. We therefore measure consistency, which quantifies how stable a model’s outputs remain as new frames arrive and penalizes unnecessary or contradictory updates.

Per-task consistency is defined as one minus a text edit–distance measure, denoted by $D$, based on the longest common substring (LCS). To account for inherent variability in the ground-truth annotations, the edit distance derived from the ground-truth captions is subtracted from that of the model outputs, as shown in \cref{eq:consistency_per_video}. Finally, per-task consistency scores are clipped between 0 and 1, and the mean across all $K$ tasks in the benchmark is reported.

\vspace{-5mm}
\begin{gather}
    \mathcal{C}_t = \frac{1}{N_t} \sum_{i=1}^{N_t-1} \left(1.0 - D\left(R_i, R_{i+1}\right) + D\left(G_i, G_{i+1}\right) \right)
    \label{eq:consistency_per_video} \tag{3} \\ 
    \mathcal{C} = \frac{1}{K} \sum_{t=1}^K \text{CLIP}(\mathcal{C}_t, 0, 1)
    \label{eq:consistency} \tag{4}
\end{gather}

\begin{table*}[h]
    \centering
    \resizebox{0.95\textwidth}{!}{
\begin{NiceTabular}{c|cccc|c|cccc|c}
    \toprule[1.5pt]
    \multirow{2}{*}{\textbf{Model}} 
        & \multicolumn{4}{c}{\textbf{Sync. Accuracy} $\uparrow$} 
        & \multirow{2}{*}{\shortstack{\textbf{Sync.}\\\textbf{Consistency} $\uparrow$}}
        & \multicolumn{4}{c}{\textbf{Async. Accuracy} $\uparrow$}
        & \multirow{2}{*}{\shortstack{\textbf{Async.}\\\textbf{Consistency} $\uparrow$}} \\
    \cmidrule(lr){2-5}\cmidrule(lr){7-10}
      & \textbf{\present{}} & \textbf{\cumulative{}} & \textbf{\future{}} & \textbf{Overall}
      &
      & \textbf{\present{}} & \textbf{\cumulative{}} & \textbf{\future{}} & \textbf{Overall}
      & \\
    \midrule[1.25pt]
    \multicolumn{11}{c}{API Models} \\
    \midrule
    \rowcolor{gray!15} GPT-5~\citep{openai2025gpt5} (low)         & \textbf{83.1} & 64.6 & 81.8 & \textbf{77.1} & 79.5 & \textcolor{gray}{34.6} & \textcolor{gray}{17.5} & \textcolor{gray}{10.1} & \textcolor{gray}{25.1} & \textcolor{gray}{99.4} \\
    GPT-5~\citep{openai2025gpt5} (medium)                         & 81.5 & \textbf{65.3} & 84.4 & \textbf{77.0} & 79.3 & \textcolor{gray}{36.3} & \textcolor{gray}{21.9} & \textcolor{gray}{12.3} & \textcolor{gray}{27.8} & \textcolor{gray}{99.6} \\
    \rowcolor{gray!15} GPT-5~\citep{openai2025gpt5} (high)        & 80.9 & 64.3 & \textbf{85.1} & 76.4 & \textbf{80.3} & \textcolor{gray}{\textbf{37.5}} & \textcolor{gray}{\textbf{26.2}} & \textcolor{gray}{\textbf{15.9}} & \textcolor{gray}{\textbf{30.3}} & \textcolor{gray}{\textbf{99.7}} \\
    \midrule
    \multicolumn{11}{c}{Open-Source Video VLMs} \\
    \midrule    
    \rowcolor{gray!15} InternVL3-2B~\citep{internvl3}  & 52.8 & 23.3 & 13.0 & 36.9 & 84.5 & 51.3 & 19.4 & 12.7 & 34.8 & 94.5 \\
    InternVL3-8B~\citep{internvl3}                     & 59.2 & 41.8 & 14.6 & 46.3 & 88.5 & 56.8 & 31.0 & 14.2 & 41.6 & 94.9 \\
    \rowcolor{gray!15} InternVL3-38B~\citep{internvl3} & 64.1 & 53.5 & 34.7 & 55.8 & \textbf{90.6} & 49.6 & 25.5 & \textbf{21.1} & 37.3 & \textbf{97.2} \\
    Qwen3-VL-2B~\citep{qwen3vl}                     & 54.3 & 38.3 & 13.1 & 42.4 & 76.4 & 53.8 & 27.1 & 9.5 & 38.0 & 96.6 \\
    \rowcolor{gray!15} Qwen3-VL-4B~\citep{qwen3vl}  & 64.3 & 47.9 & 19.5 & 51.6 & 84.1 & \textbf{62.3} & \textbf{36.4} & 17.9 & \textbf{46.8} & 95.3 \\
    Qwen3-VL-8B~\citep{qwen3vl}                     & 61.7 & 52.3 & \textbf{35.5} & 54.3 & 86.0 & 57.9 & 34.3 & \textbf{20.7} & 44.3 & 96.0 \\
    \rowcolor{gray!15} Qwen3-VL-32B~\citep{qwen3vl} & \textbf{69.4} & \textbf{58.3} & 31.4 & \textbf{59.5} & 84.9 & 54.8 & 32.1 & 14.5 & 41.0 & 97.1 \\
    \midrule
    \multicolumn{11}{c}{Open-Source Streaming VLMs} \\
    \midrule
    \rowcolor{gray!15} VideoLLM-Online~\citep{chen2024videollm}    & 20.0 & 1.6  & 2.3  & 11.3 & 74.9 & 12.9 & 1.9 & 3.4 & 7.9 & \textbf{97.5} \\
    FlashVStream~\citep{zhang2024flash}                          & 19.8 & 5.6  & 8.7  & 13.5 & 86.8 & 22.2 & 10.0 & 10.5 & 16.5 & 89.9 \\
    \rowcolor{gray!15} Dispider~\citep{qian2025dispider}           & 48.5 & \textbf{33.5} & \textbf{66.6} & \textbf{46.8} & \textbf{94.2} & \textbf{49.9} & \textbf{27.5} & \textbf{53.1} & \textbf{43.4} & \textbf{97.5} \\
    StreamBridge~\citep{wang2025streambridge}                          & \textbf{50.1} & 26.8 & 14.1 & 36.8 & 85.0 & 34.2 & 6.9 & 10.3 & 21.7 & 94.0 \\
    \bottomrule[1.5pt] 
\end{NiceTabular}
}\label{tab:main_results}

    \caption{\textbf{Streaming-adapted video VLMs surpass prior streaming VLMs on \ours{}.}
     Models are grouped by their streaming capability and being open- or closed-source. Hardware, model configurations, and protocol settings are detailed in \cref{sec:results}. Video models are adapted for asynchronous evaluation using the method in \cref{sec:video_to_streaming} with \SWaccess{} policy. We highlight the best numbers in each column per model category and for accuracies, we highlight numbers within maximum standard deviation of judge evaluations across tasks (0.6\%). GPT-5 is accessed via a web API, and the resulting network latency hurts performance under the asynchronous protocol relative to locally deployed models such as Qwen3-VL and InternVL3; we therefore gray out these results in the table.}\label{tab:main_results}
    \vspace{-2mm}
\end{table*}

\begin{table}[h]
\centering
\resizebox{0.75\columnwidth}{!}{%
\begin{tabular}{lcccc}
\toprule
\multirow{2}{*}{\vspace{-0.15cm}\shortstack[l]{\textbf{Memory}\\\textbf{Policy}}} & \multicolumn{3}{c}{\textbf{Tasks}} & \multirow{2}{*}{\textbf{Overall}} \\
\cmidrule(lr){2-4}
  & \textbf{\present{}} & \textbf{\cumulative{}} & \textbf{\future{}} &  \\
\midrule
\multicolumn{5}{c}{\textbf{GPT-5}} \\
\midrule
\rowcolor{gray!15} SW   & 81.5 & 65.3 & 84.4 & 77.0 \\
U                       & 81.1 & \textbf{72.3} & 82.1 & 78.5 \\
\rowcolor{gray!15} SW+U & \textbf{83.7} & \textbf{72.6} & \textbf{85.7} & \textbf{80.6} \\

\midrule
\multicolumn{5}{c}{\textbf{Qwen3-VL-8B}} \\
\midrule
\rowcolor{gray!15} SW   & 61.7 & 52.3 & \textbf{35.5} & 54.3 \\
U                       & \textbf{62.6} & \textbf{52.8} & 34.8 & \textbf{54.9} \\
\rowcolor{gray!15} SW+U & \textbf{62.4} & \textbf{52.6} & \textbf{35.1} & \textbf{54.8} \\

\bottomrule
\end{tabular}
}
\caption{\textbf{Effect of memory policies.} \SWUaccess{} performs best or within one standard deviation of the judge (0.6\%). We use \SWaccess{} in other experiments as the only feasible policy for long videos. We report under synchronous protocol with maximum context size of 64 frames. We highlight the best numbers per model.}
\label{tab:abl_memory_management}
\vspace{-8px}
\end{table}

\subsubsection{LLM-Based Similarity Evaluation}\label{sec:llm_judge}

We employ GPT-5 (with medium reasoning) as an LLM-based judge to assess both syntactic and semantic similarity between a model’s generated response $R_i$ and its corresponding ground-truth annotation $G_i$ at timestep $i$. Our evaluation framework follows the general principles of G-Eval~\citep{geval} but differs in that each judgment is made relative to a reference annotation rather than in isolation. We rely on the publicly available OpenAI GPT-5 (medium reasoning) model via the official API without any fine-tuning, and include a detailed evaluation guidelines within the judge prompt to ensure consistent and reproducible scoring across task types. For each $(G_i, R_i)$ pair, the judge outputs both a binary score (${ \text{yes}|\text{no} }$) and a rubric score in $\{0, 1, 2, 3\}$, where higher values indicate stronger alignment. Empirically, both the binary and rubric scores exhibit low variance with less than 0.6\% standard deviation across five independent judge evaluations. We find the binary metric sufficient to capture performance gaps between models and evaluation protocols; hence, all reported accuracy results are based on the binary metric. In addition, we validated the agreement between GPT-Judge and human experts with a held-out set of 450 samples (across a mixture of task types) which yielded a Cohen’s ${\kappa}$ of 0.91 indicating very high agreement. More details regarding the exact judge prompts we used can be found in the Appendix.

\section{Evaluation} \label{sec:eval}

\subsection{Turning Video VLMs to Streaming VLMs}\label{sec:video_to_streaming}
Conventional video VLMs lack native support for evaluation under asynchronous protocol. To enable streaming-capable behavior, we introduce two external components: a model \textit{memory buffer} separate from the camera buffer that stores either input frames or other intermediate representations and a \textit{working context} that is selected from memory buffer with relative information for inference.

\textbf{Memory policy}.
A memory policy decides
(1) what to keep in the memory buffer when retrieving new frames and
(2) how to select working context from memory buffer for each inference step.
Under the asynchronous protocol, incoming frames are added to the model’s memory buffer. At each inference step, the model queries this buffer to assemble its working context, governed by the memory policy and the maximum context size $k$ (the maximum number of frames to include in the context).

We consider three different policies: Sliding Window (\SWaccess{}), Uniform (\Uaccess{}), and Sliding Window with a Uniform tail (\SWUaccess{}). Sliding Window (\SWaccess{}) policy stores only  the latest $k$ frames in the memory buffer and the working context selects all frames in the memory buffer. We use \SWaccess{} as the default policy since it is the only feasible policy for long videos. For the Uniform (\Uaccess{}) policy, the memory buffer keeps all frames and the working context uniformly samples $k$ frames for inference. For the Sliding Window with a Uniform tail (\SWUaccess{}) policy, the memory buffer also keeps all frames while the working context selects the most recent $k/2$ frames plus $k/2$ uniformly sampled older frames, balancing recency and temporal coverage.

We compare the memory policies under the asynchronous protocol in \cref{tab:abl_memory_management}, using a maximum context size of $k=64$. While \SWaccess{} favors recency, we find that giving up some recency for broader temporal coverage can help. The \Uaccess{} policy brings both early and recent video frames into context and improves \cumulative{} tasks, showing that long-range information is useful when past events matter. The hybrid \SWUaccess{} policy offers a balanced trade-off but yields only marginal gains with Qwen3-VL-8B, which may depend on tuning the tail size. Note that with our default policy (\SWaccess{}), the memory buffer size equals $k$, so we refer to it simply as the model buffer size throughout the paper.

\subsection{Main Results}\label{sec:results}
\noindent \textbf{Setup.}
All models were evaluated on identical infrastructure equipped with NVIDIA H100 GPUs. Unless mentioned otherwise, smaller models (2B and 4B) were run on a single GPU, medium models (8B) used two and larger variants (32B and 38B) utilized four GPUs per run. All Streaming VLMs were run on a single GPU. For all models, we use CUDA 12.4 with FlashAttention v2.7.3 and set the computation precision to bfloat16 to ensure fairness in evaluation. Video VLMs are modified as described in \cref{sec:video_to_streaming} to enable their execution under the asynchronous protocol. For all the results presented in \cref{tab:main_results}, we set the camera buffer size to 600 and memory buffer size to 64. A detailed ablation on camera buffer size will be presented in Appendix.

\noindent \textbf{Model Comparison.}
In \cref{tab:main_results}, we report the mean average accuracy, $\mathcal{A}$, and mean average consistency, $\mathcal{C}$, of all evaluated models under both synchronous and asynchronous evaluation protocols. We further provide a breakdown of accuracy across the three primary task types.
In the synchronous setting, we observe the typical trend consistent with most evaluation benchmarks: larger and more capable models achieve higher performance. This pattern holds both within model families, for example, Qwen3-VL 8B outperforms its smaller 2B counterpart and across model categories, where a presumably substantially larger model such as GPT-5 surpasses Qwen3-VL 32B. 

Under the asynchronous protocol, which represents the true streaming scenario, we observe that dedicated streaming models offer limited advantages over our streaming-adapted video VLMs. Notably, smaller models such as Qwen3-VL-4B outperform several recent streaming architectures.
Streaming-adapted Qwen3-VL-4B surpasses Dispider, the best streaming VLM, by 3\%. We also observe a 16\% gap between GPT-5 and Qwen3-VL-4B based on latency and API response times at the time of evaluation. Amongst streaming models, Dispider demonstrates a competitive performance, particularly on the \future{} task. But the overall underperformance of streaming VLMs may stem from their fine-tuning on narrow, task-specific domains, whereas our streaming-adapted video VLMs are training-free extensions of the base models which preserve broader generalization capabilities. 

In \cref{tab:main_results}, we report mean average consistency under synchronous and asynchronous protocols. In the synchronous setting, streaming VLMs outperform both video models and the larger GPT-5, reflecting their training to maintain stable outputs even when accuracy is lower. Video models, while often more accurate, tend to be more verbose and produce varied phrasings across frames, which manifests as inconsistency. Since consistency is measured at the camera sampling rate, and models respond less often under the asynchronous protocol, consistency scores rise across the board. In this regime, the gap between streaming models and streaming-adapted video models narrows substantially.

\noindent\textbf{Evaluating API Models.} Closed-source API models, such as GPT-5, are evaluated under the same sync. and async. protocols as other VLMs. At each step, we query the model (\texttt{gpt-5-2025-08-07}) via the \texttt{openai} package, to submit the task prompt and the selected frames returned by the memory policy. Under the async. protocol, accuracy reflects both model inference latency and network latency.

\begin{table}[t]
\centering
\resizebox{0.95\columnwidth}{!}{%
\begin{tabular}{llcccc}
\toprule
\multirow{2}{*}{\textbf{Model}} & \multirow{2}{*}{\textbf{Accel.}} & \multicolumn{3}{c}{\textbf{Tasks}} & \multirow{2}{*}{\textbf{Overall}}\\
\cmidrule(lr){3-5}
 &  & \textbf{\present{}} & \textbf{\cumulative{}} & \textbf{\future{}} &  \\
\midrule
\multirow{2}{*}{Qwen3-VL-2B}  & \cellcolor{gray!15}A100 & \cellcolor{gray!15}52.1 & \cellcolor{gray!15}22.8 &  \cellcolor{gray!15}9.4 & \cellcolor{gray!15}35.8 \\
                              & H100 & 53.8 & 27.1 &  9.5 & 38.0 \\
\addlinespace[0.3em]
\multirow{2}{*}{Qwen3-VL-4B}  & \cellcolor{gray!15}A100 & \cellcolor{gray!15}58.9 & \cellcolor{gray!15}31.9 & \cellcolor{gray!15}16.3 & \cellcolor{gray!15}43.3 \\
                              & H100 & 62.3 & 36.4 & 17.9 & 46.8 \\
\addlinespace[0.3em]
\multirow{2}{*}{Qwen3-VL-8B}  & \cellcolor{gray!15}A100 & \cellcolor{gray!15}55.3 & \cellcolor{gray!15}32.5 & \cellcolor{gray!15}16.8 & \cellcolor{gray!15}41.7 \\
                              & H100 & 57.9 & 34.3 & 20.7 & 44.3 \\
\addlinespace[0.3em]
\multirow{2}{*}{Qwen3-VL-32B} & \cellcolor{gray!15}A100 & \cellcolor{gray!15}51.0 & \cellcolor{gray!15}27.0 & \cellcolor{gray!15}11.0 & \cellcolor{gray!15}36.8 \\
                              & H100 & 54.8 & 32.1 & 14.5 & 41.0 \\
\bottomrule
\end{tabular}
}
\vspace*{-3pt}
\caption{\textbf{Effect of Hardware Accelerators.} We evaluate Qwen3-VL model family using the asynchronous protocol on Nvidia A100 and H100. All runs use memory buffer size 64 and \SWaccess{} policy. 
}
\label{tab:async_accel_compare}
\vspace{-3mm}
\end{table}

\begin{table}[h]
\centering
\resizebox{0.80\columnwidth}{!}{%
\begin{tabular}{ccccc}
\toprule
 \multirow{2}{*}{\vspace{-0.2cm}\shortstack[c]{\textbf{Memory}\\\textbf{Buffer Size}}} & \multicolumn{3}{c}{\textbf{Tasks}} & \multirow{2}{*}{\textbf{Overall}} \\
\cmidrule(lr){2-4}
 & \textbf{\present{}} & \textbf{\cumulative{}} & \textbf{\future{}} &  \\
\midrule
\rowcolor{gray!15} 1    & 42.0 & 15.8 & \textbf{31.0} & 42.0 \\
8    & \textbf{66.1} & 24.5 & 30.0 & \textbf{47.0} \\
\rowcolor{gray!15} 16   & 63.4 & 30.0 & 26.5 & 46.8 \\
32   & 60.1 & \textbf{35.3} & 23.0 & 46.1 \\
\rowcolor{gray!15} 64   & 57.9 & 34.3 & 20.7 & 44.3 \\
\bottomrule
\end{tabular}
}
\vspace*{-3pt}
\caption{\textbf{Effect of memory buffer size.}  We evaluate Qwen3-VL-8B model using the asynchronous protocol. Image resolution is fixed across all buffer sizes, and all runs use the \SWaccess{} policy. We highlight the best numbers in each column.}
\label{tab:mem_buffer_size}
\vspace{-3mm}
\end{table}

\begin{table}[h]
\centering
\resizebox{0.78\columnwidth}{!}{%
\begin{tabular}{ccccc}
\toprule
 \multirow{2}{*}{\vspace{-0.2cm}\shortstack[c]{\textbf{Min}\\\textbf{Pixels}}} & \multicolumn{3}{c}{\textbf{Tasks}} & \multirow{2}{*}{\textbf{Overall}}  \\
\cmidrule(lr){2-4}
 & \textbf{\present{}} & \textbf{\cumulative{}} & \textbf{\future{}} &  \\
\midrule
 \rowcolor{gray!15} 16,384  & 45.0 & 29.3 & 17.8 & 35.5 \\
 50,176  & 55.5 & 28.5 & \textbf{21.9} & 41.4 \\
 \rowcolor{gray!15} 200,704 & 57.9 & 34.3 & 20.7 & 44.3 \\
 262,144 & \textbf{59.8} & \textbf{38.0} & 19.9 & \textbf{46.3} \\
 \rowcolor{gray!15} 409,600 & 58.5 & 36.6 & 18.6 & 45.0 \\
\bottomrule
\end{tabular}
}
\vspace*{-3pt}
\caption{\textbf{Effect of input resolution.} We evaluate Qwen3-VL-8B using the asynchronous protocol. All runs use memory buffer size 64 and \SWaccess{} policy. We highlight the best numbers in each column.}
\label{tab:img_resolution}
\vspace{-3mm}
\end{table}

\subsection{Ablations using Asynchronous Protocol}\label{sec:async_ablations}
The asynchronous protocol enables systematic benchmarking and hardware-aware tuning, allowing models to be configured for optimal real-time performance.

\noindent
\textbf{On-device Compute.}
Our framework enables systematic comparison of different hardware accelerators for a given model. As shown in \cref{tab:async_accel_compare}, variations in underlying hardware can significantly impact model accuracy. This evaluation can be naturally extended to edge devices, allowing developers to make informed trade-offs among available compute substrates when deploying VLMs for streaming.

\noindent
\textbf{Memory buffer size.}
Our benchmark comprises a balanced mix of short- and long-horizon tasks, enabling systematic analysis of memory buffer configurations under the asynchronous protocol. As shown in \Cref{tab:mem_buffer_size}, varying buffer size exposes a clear trade-off between temporal context and latency: larger buffers provide richer context but incur higher delays. Unlike prior benchmarks, ours explicitly captures this interplay between context length and inference latency.

\noindent
\textbf{Image Resolution.}
A similar trade-off between spatial resolution and latency is observed in \Cref{tab:img_resolution}. Increasing spatial resolution captures finer details but can degrade performance under the tight latency limits of streaming applications, as reflected in our asynchronous evaluation protocol.

\subsection{Model-specific Optimizations}\label{sec:other_optimizations}
To improve VLMs performance in a streaming setup like \ours{}, we recommend several model-specifc optimizations beyond the choices discussed above: (1) prompt engineering to keep responses concise, (2) reusing previously computed visual tokens for the frames reappearing in memory buffer, (3) reusing the LLM KV-cache to cut prefill time, and (4) reusing previously generated tokens, for example via speculative decoding~\citep{leviathan2023fast}, since earlier outputs may still be (partially) valid, such as in OCR cases.

\section{Conclusion}
We introduced \ours{}, an evaluation framework for assessing streaming VLMs under realistic, time-constrained settings. \ours{} features temporally dense annotations across diverse domains and tasks with synchronous and asynchronous evaluation protocols that capture the interplay between accuracy and latency, factors overlooked by prior benchmarks. Through extensive experiments, we revealed trade-offs between computational efficiency and contextual visual reasoning. Finally, we proposed a training-free method to adapt video VLMs for streaming, achieving superior generalization compared to recent streaming VLMs. We hope \ours{} serves as a foundation for developing efficient, latency-aware,  interactive multimodal systems.
{
    \small
    \bibliographystyle{ieeenat_fullname}
    \bibliography{main}

\begin{thebibliography}{36}
\providecommand{\natexlab}[1]{#1}
\providecommand{\url}[1]{\texttt{#1}}
\expandafter\ifx\csname urlstyle\endcsname\relax
  \providecommand{\doi}[1]{doi: #1}\else
  \providecommand{\doi}{doi: \begingroup \urlstyle{rm}\Url}\fi

\bibitem[Bai et~al.(2025)Bai, Chen, Liu, Wang, Ge, Song, Dang, Wang, Wang, Tang, Zhong, Zhu, Yang, Li, Wan, Wang, Ding, Fu, Xu, Ye, Zhang, Xie, Cheng, Zhang, Yang, Xu, and Lin]{Qwen2.5-VL}
Shuai Bai, Keqin Chen, Xuejing Liu, Jialin Wang, Wenbin Ge, Sibo Song, Kai Dang, Peng Wang, Shijie Wang, Jun Tang, Humen Zhong, Yuanzhi Zhu, Mingkun Yang, Zhaohai Li, Jianqiang Wan, Pengfei Wang, Wei Ding, Zheren Fu, Yiheng Xu, Jiabo Ye, Xi Zhang, Tianbao Xie, Zesen Cheng, Hang Zhang, Zhibo Yang, Haiyang Xu, and Junyang Lin.
\newblock Qwen2.5-vl technical report.
\newblock \emph{arXiv preprint arXiv:2502.13923}, 2025.

\bibitem[Caba~Heilbron et~al.(2015)Caba~Heilbron, Escorcia, Ghanem, and Carlos~Niebles]{caba2015activitynet}
Fabian Caba~Heilbron, Victor Escorcia, Bernard Ghanem, and Juan Carlos~Niebles.
\newblock Activitynet: A large-scale video benchmark for human activity understanding.
\newblock In \emph{Proceedings of the ieee conference on computer vision and pattern recognition}, pages 961--970, 2015.

\bibitem[Cai et~al.(2024)Cai, Tan, Zhang, Zou, Zhang, Yao, Zhu, Gu, Zhong, Shang, Dou, Park, Gao, Lee, and Yang]{cai2024temporalbenchbenchmarkingfinegrainedtemporal}
Mu Cai, Reuben Tan, Jianrui Zhang, Bocheng Zou, Kai Zhang, Feng Yao, Fangrui Zhu, Jing Gu, Yiwu Zhong, Yuzhang Shang, Yao Dou, Jaden Park, Jianfeng Gao, Yong~Jae Lee, and Jianwei Yang.
\newblock Temporalbench: Benchmarking fine-grained temporal understanding for multimodal video models, 2024.

\bibitem[Chen et~al.(2024)Chen, Lv, Wu, Lin, Song, Gao, Liu, Gao, Mao, and Shou]{chen2024videollm}
Joya Chen, Zhaoyang Lv, Shiwei Wu, Kevin~Qinghong Lin, Chenan Song, Difei Gao, Jia-Wei Liu, Ziteng Gao, Dongxing Mao, and Mike~Zheng Shou.
\newblock Videollm-online: Online video large language model for streaming video.
\newblock In \emph{Proceedings of the IEEE/CVF Conference on Computer Vision and Pattern Recognition}, pages 18407--18418, 2024.

\bibitem[Chen et~al.(2025)Chen, Zeng, Lin, Li, Ma, and Shou]{livecc}
Joya Chen, Ziyun Zeng, Yiqi Lin, Wei Li, Zejun Ma, and Mike~Zheng Shou.
\newblock Livecc: Learning video llm with streaming speech transcription at scale.
\newblock In \emph{CVPR}, 2025.

\bibitem[Comanici et~al.(2025)Comanici, Bieber, Schaekermann, Pasupat, Sachdeva, Dhillon, Blistein, Ram, Zhang, Rosen, et~al.]{comanici2025gemini}
Gheorghe Comanici, Eric Bieber, Mike Schaekermann, Ice Pasupat, Noveen Sachdeva, Inderjit Dhillon, Marcel Blistein, Ori Ram, Dan Zhang, Evan Rosen, et~al.
\newblock Gemini 2.5: Pushing the frontier with advanced reasoning, multimodality, long context, and next generation agentic capabilities.
\newblock \emph{arXiv preprint arXiv:2507.06261}, 2025.

\bibitem[Fu et~al.(2024)Fu, Dai, Luo, Li, Ren, Zhang, Wang, Zhou, Shen, Zhang, et~al.]{videomme}
Chaoyou Fu, Yuhan Dai, Yondong Luo, Lei Li, Shuhuai Ren, Renrui Zhang, Zihan Wang, Chenyu Zhou, Yunhang Shen, Mengdan Zhang, et~al.
\newblock Video-mme: The first-ever comprehensive evaluation benchmark of multi-modal llms in video analysis.
\newblock \emph{arXiv preprint arXiv:2405.21075}, 2024.

\bibitem[Grauman et~al.(2022)Grauman, Westbury, Byrne, Chavis, Furnari, Girdhar, Hamburger, Jiang, Liu, Liu, et~al.]{grauman2022ego4d}
Kristen Grauman, Andrew Westbury, Eugene Byrne, Zachary Chavis, Antonino Furnari, Rohit Girdhar, Jackson Hamburger, Hao Jiang, Miao Liu, Xingyu Liu, et~al.
\newblock Ego4d: Around the world in 3,000 hours of egocentric video.
\newblock In \emph{Proceedings of the IEEE/CVF conference on computer vision and pattern recognition}, pages 18995--19012, 2022.

\bibitem[Hong et~al.(2024)Hong, Wang, Ding, Yu, Lv, Wang, Cheng, Huang, Ji, Xue, et~al.]{hong2024cogvlm2}
Wenyi Hong, Weihan Wang, Ming Ding, Wenmeng Yu, Qingsong Lv, Yan Wang, Yean Cheng, Shiyu Huang, Junhui Ji, Zhao Xue, et~al.
\newblock Cogvlm2: Visual language models for image and video understanding.
\newblock \emph{arXiv preprint arXiv:2408.16500}, 2024.

\bibitem[Hurst et~al.(2024)Hurst, Lerer, Goucher, Perelman, Ramesh, Clark, Ostrow, Welihinda, Hayes, Radford, et~al.]{hurst2024gpt}
Aaron Hurst, Adam Lerer, Adam~P Goucher, Adam Perelman, Aditya Ramesh, Aidan Clark, AJ Ostrow, Akila Welihinda, Alan Hayes, Alec Radford, et~al.
\newblock Gpt-4o system card.
\newblock \emph{arXiv preprint arXiv:2410.21276}, 2024.

\bibitem[Leviathan et~al.(2023)Leviathan, Kalman, and Matias]{leviathan2023fast}
Yaniv Leviathan, Matan Kalman, and Yossi Matias.
\newblock Fast inference from transformers via speculative decoding.
\newblock In \emph{International Conference on Machine Learning}, pages 19274--19286. PMLR, 2023.

\bibitem[Li et~al.(2024)Li, Zhang, Guo, Zhang, Li, Zhang, Zhang, Li, Liu, and Li]{llava-onevision}
Bo Li, Yuanhan Zhang, Dong Guo, Renrui Zhang, Feng Li, Hao Zhang, Kaichen Zhang, Yanwei Li, Ziwei Liu, and Chunyuan Li.
\newblock Llava-onevision: Easy visual task transfer, 2024.

\bibitem[Lin et~al.(2024)Lin, Fang, Chen, Wan, Luo, Li, Liu, and Sun]{lin2024streamingbench}
Junming Lin, Zheng Fang, Chi Chen, Zihao Wan, Fuwen Luo, Peng Li, Yang Liu, and Maosong Sun.
\newblock Streamingbench: Assessing the gap for mllms to achieve streaming video understanding.
\newblock \emph{arXiv preprint arXiv:2411.03628}, 2024.

\bibitem[Liu et~al.(2023)Liu, Iter, Xu, Wang, Xu, and Zhu]{geval}
Yang Liu, Dan Iter, Yichong Xu, Shuohang Wang, Ruochen Xu, and Chenguang Zhu.
\newblock {G}-eval: {NLG} evaluation using gpt-4 with better human alignment.
\newblock In \emph{Proceedings of the 2023 Conference on Empirical Methods in Natural Language Processing}, pages 2511--2522, Singapore, 2023. Association for Computational Linguistics.

\bibitem[Liu et~al.(2024{\natexlab{a}})Liu, Ma, Qi, Wu, Chen, and Shan]{liu2024etbench}
Ye Liu, Zongyang Ma, Zhongang Qi, Yang Wu, Chang~Wen Chen, and Ying Shan.
\newblock E.t. bench: Towards open-ended event-level video-language understanding.
\newblock In \emph{Neural Information Processing Systems (NeurIPS)}, 2024{\natexlab{a}}.

\bibitem[Liu et~al.(2024{\natexlab{b}})Liu, Zhu, Shi, Zhang, Lou, Yang, Xi, Cao, Gu, Li, Li, Fang, Chen, Hsieh, Huang, Cheng, Nath, Hu, Liu, Krishna, Xu, Wang, Molchanov, Kautz, Yin, Han, and Lu]{liu2024nvila}
Zhijian Liu, Ligeng Zhu, Baifeng Shi, Zhuoyang Zhang, Yuming Lou, Shang Yang, Haocheng Xi, Shiyi Cao, Yuxian Gu, Dacheng Li, Xiuyu Li, Yunhao Fang, Yukang Chen, Cheng-Yu Hsieh, De-An Huang, An-Chieh Cheng, Vishwesh Nath, Jinyi Hu, Sifei Liu, Ranjay Krishna, Daguang Xu, Xiaolong Wang, Pavlo Molchanov, Jan Kautz, Hongxu Yin, Song Han, and Yao Lu.
\newblock Nvila: Efficient frontier visual language models, 2024{\natexlab{b}}.

\bibitem[Mangalam et~al.(2023)Mangalam, Akshulakov, and Malik]{mangalam2023egoschema}
Karttikeya Mangalam, Raiymbek Akshulakov, and Jitendra Malik.
\newblock Egoschema: A diagnostic benchmark for very long-form video language understanding, 2023.

\bibitem[Marafioti et~al.(2025)Marafioti, Zohar, Farré, Noyan, Bakouch, Cuenca, Zakka, Allal, Lozhkov, Tazi, Srivastav, Lochner, Larcher, Morlon, Tunstall, von Werra, and Wolf]{marafioti2025smolvlm}
Andrés Marafioti, Orr Zohar, Miquel Farré, Merve Noyan, Elie Bakouch, Pedro Cuenca, Cyril Zakka, Loubna~Ben Allal, Anton Lozhkov, Nouamane Tazi, Vaibhav Srivastav, Joshua Lochner, Hugo Larcher, Mathieu Morlon, Lewis Tunstall, Leandro von Werra, and Thomas Wolf.
\newblock Smolvlm: Redefining small and efficient multimodal models.
\newblock \emph{arXiv preprint arXiv:2504.05299}, 2025.

\bibitem[Niu et~al.(2025)Niu, Li, Miao, Ge, Zhou, He, Dong, Duan, Ding, Qian, et~al.]{niu2025ovo}
Junbo Niu, Yifei Li, Ziyang Miao, Chunjiang Ge, Yuanhang Zhou, Qihao He, Xiaoyi Dong, Haodong Duan, Shuangrui Ding, Rui Qian, et~al.
\newblock Ovo-bench: How far is your video-llms from real-world online video understanding?
\newblock In \emph{Proceedings of the Computer Vision and Pattern Recognition Conference}, pages 18902--18913, 2025.

\bibitem[{OpenAI}(2025)]{openai2025gpt5}
{OpenAI}.
\newblock Gpt-5 system card.
\newblock Technical Report TR-GPT5-2025, OpenAI, 2025.
\newblock Technical report. Available at \url{https://cdn.openai.com/gpt-5-system-card.pdf} (accessed 2025-11-13).

\bibitem[Pătrăucean et~al.(2023)Pătrăucean, Smaira, Gupta, Continente, Markeeva, Banarse, Koppula, Heyward, Malinowski, Yang, Doersch, Matejovicova, Sulsky, Miech, Frechette, Klimczak, Koster, Zhang, Winkler, Aytar, Osindero, Damen, Zisserman, and Carreira]{patraucean2023perceptiontest}
Viorica Pătrăucean, Lucas Smaira, Ankush Gupta, Adrià~Recasens Continente, Larisa Markeeva, Dylan Banarse, Skanda Koppula, Joseph Heyward, Mateusz Malinowski, Yi Yang, Carl Doersch, Tatiana Matejovicova, Yury Sulsky, Antoine Miech, Alex Frechette, Hanna Klimczak, Raphael Koster, Junlin Zhang, Stephanie Winkler, Yusuf Aytar, Simon Osindero, Dima Damen, Andrew Zisserman, and João Carreira.
\newblock Perception test: A diagnostic benchmark for multimodal video models.
\newblock In \emph{Advances in Neural Information Processing Systems}, 2023.

\bibitem[Qian et~al.(2025)Qian, Ding, Dong, Zhang, Zang, Cao, Lin, and Wang]{qian2025dispider}
Rui Qian, Shuangrui Ding, Xiaoyi Dong, Pan Zhang, Yuhang Zang, Yuhang Cao, Dahua Lin, and Jiaqi Wang.
\newblock Dispider: Enabling video llms with active real-time interaction via disentangled perception, decision, and reaction.
\newblock In \emph{Proceedings of the Computer Vision and Pattern Recognition Conference}, pages 24045--24055, 2025.

\bibitem[Tang et~al.(2019)Tang, Ding, Rao, Zheng, Zhang, Zhao, Lu, and Zhou]{tang2019coin}
Yansong Tang, Dajun Ding, Yongming Rao, Yu Zheng, Danyang Zhang, Lili Zhao, Jiwen Lu, and Jie Zhou.
\newblock Coin: A large-scale dataset for comprehensive instructional video analysis.
\newblock In \emph{Proceedings of the IEEE/CVF Conference on Computer Vision and Pattern Recognition}, pages 1207--1216, 2019.

\bibitem[Team(2025)]{qwen3vl}
Qwen Team.
\newblock Qwen3 technical report, 2025.

\bibitem[Vasu et~al.(2025)Vasu, Faghri, Li, Koc, True, Antony, Santhanam, Gabriel, Grasch, Tuzel, and Pouransari]{fastvlm2025}
Pavan Kumar~Anasosalu Vasu, Fartash Faghri, Chun-Liang Li, Cem Koc, Nate True, Albert Antony, Gokul Santhanam, James Gabriel, Peter Grasch, Oncel Tuzel, and Hadi Pouransari.
\newblock Fastvlm: Efficient vision encoding for vision language models.
\newblock In \emph{Proceedings of the IEEE/CVF Conference on Computer Vision and Pattern Recognition (CVPR)}, 2025.

\bibitem[Wang et~al.(2025)Wang, Feng, Lai, Xu, Li, Ge, Dehghan, Cao, and Huang]{wang2025streambridge}
Haibo Wang, Bo Feng, Zhengfeng Lai, Mingze Xu, Shiyu Li, Weifeng Ge, Afshin Dehghan, Meng Cao, and Ping Huang.
\newblock Streambridge: Turning your offline video large language model into a proactive streaming assistant.
\newblock \emph{arXiv preprint arXiv:2505.05467}, 2025.

\bibitem[Wang et~al.(2024{\natexlab{a}})Wang, Bai, Tan, Wang, Fan, Bai, Chen, Liu, Wang, Ge, Fan, Dang, Du, Ren, Men, Liu, Zhou, Zhou, and Lin]{Qwen2-VL}
Peng Wang, Shuai Bai, Sinan Tan, Shijie Wang, Zhihao Fan, Jinze Bai, Keqin Chen, Xuejing Liu, Jialin Wang, Wenbin Ge, Yang Fan, Kai Dang, Mengfei Du, Xuancheng Ren, Rui Men, Dayiheng Liu, Chang Zhou, Jingren Zhou, and Junyang Lin.
\newblock Qwen2-vl: Enhancing vision-language model's perception of the world at any resolution.
\newblock \emph{arXiv preprint arXiv:2409.12191}, 2024{\natexlab{a}}.

\bibitem[Wang et~al.(2024{\natexlab{b}})Wang, Chen, Wang, Cao, Liu, Gao, Zhu, Zhu, Lu, Qiao, and Dai]{internvl3}
Weiyun Wang, Zhe Chen, Wenhai Wang, Yue Cao, Yangzhou Liu, Zhangwei Gao, Jinguo Zhu, Xizhou Zhu, Lewei Lu, Yu Qiao, and Jifeng Dai.
\newblock Enhancing the reasoning ability of multimodal large language models via mixed preference optimization.
\newblock \emph{arXiv preprint arXiv:2411.10442}, 2024{\natexlab{b}}.

\bibitem[Wu et~al.(2021)Wu, Yu, Chen, Tenenbaum, and Gan]{wu2021star_situated_reasoning}
Bo Wu, Shoubin Yu, Zhenfang Chen, Joshua~B Tenenbaum, and Chuang Gan.
\newblock Star: A benchmark for situated reasoning in real-world videos.
\newblock In \emph{Thirty-fifth Conference on Neural Information Processing Systems (NeurIPS)}, 2021.

\bibitem[Wu et~al.(2024)Wu, Li, Chen, and Li]{wu2024longvideobench}
Haoning Wu, Dongxu Li, Bei Chen, and Junnan Li.
\newblock Longvideobench: A benchmark for long-context interleaved video-language understanding, 2024.

\bibitem[Xiao et~al.(2021)Xiao, Shang, Yao, and Chua]{xiao2021nextqa}
Junbin Xiao, Xindi Shang, Angela Yao, and Tat-Seng Chua.
\newblock Next-qa: Next phase of question-answering to explaining temporal actions.
\newblock In \emph{Proceedings of the IEEE/CVF Conference on Computer Vision and Pattern Recognition (CVPR)}, pages 9777--9786, 2021.

\bibitem[Xu et~al.(2016)Xu, Mei, Yao, and Rui]{xu2016msrvtt}
Jun Xu, Tao Mei, Ting Yao, and Yong Rui.
\newblock Msr-vtt: A large video description dataset for bridging video and language.
\newblock In \emph{Proceedings of the IEEE conference on computer vision and pattern recognition}, pages 5288--5296, 2016.

\bibitem[Xu et~al.(2025)Xu, Xiao, Chen, He, Peng, Lu, and Han]{xu2025streamingvlm}
Ruyi Xu, Guangxuan Xiao, Yukang Chen, Liuning He, Kelly Peng, Yao Lu, and Song Han.
\newblock Streamingvlm: Real-time understanding for infinite video streams.
\newblock \emph{arXiv preprint arXiv:2510.09608}, 2025.

\bibitem[Yao et~al.(2024)Yao, Yu, Zhang, Wang, Cui, Zhu, Cai, Li, Zhao, He, et~al.]{yao2024minicpm}
Yuan Yao, Tianyu Yu, Ao Zhang, Chongyi Wang, Junbo Cui, Hongji Zhu, Tianchi Cai, Haoyu Li, Weilin Zhao, Zhihui He, et~al.
\newblock Minicpm-v: A gpt-4v level mllm on your phone.
\newblock \emph{arXiv preprint arXiv:2408.01800}, 2024.

\bibitem[Zhang et~al.(2024)Zhang, Wang, Tang, Liu, Feng, Dai, and Jin]{zhang2024flash}
Haoji Zhang, Yiqin Wang, Yansong Tang, Yong Liu, Jiashi Feng, Jifeng Dai, and Xiaojie Jin.
\newblock Flash-vstream: Memory-based real-time understanding for long video streams.
\newblock \emph{arXiv preprint arXiv:2406.08085}, 2024.

\bibitem[Zhou et~al.(2018)Zhou, Xu, and Corso]{youcook2}
Luowei Zhou, Chenliang Xu, and Jason Corso.
\newblock Towards automatic learning of procedures from web instructional videos.
\newblock In \emph{Proceedings of the AAAI Conference on Artificial Intelligence}, 2018.

\end{thebibliography}
}

\clearpage
\setcounter{page}{1}
\maketitlesupplementary
\appendix

\section{Dataset Preprocessing and Annotations}
As detailed in \cref{sec:dataset}, our benchmark comprises 44 videos sourced from existing datasets and 48 newly recorded videos. For all newly recorded content, we apply strict anonymization to remove personally identifiable information. Specifically, we run a face detector on every frame and blur all detected faces, using a threshold tuned for high recall to minimize missed detections. We additionally blur all visible vehicle license plates. No further personal data from human subjects is collected.

\begin{figure*}[t]
    \centering
    \includegraphics[width=0.95\textwidth, height=0.93\textheight, keepaspectratio=true]{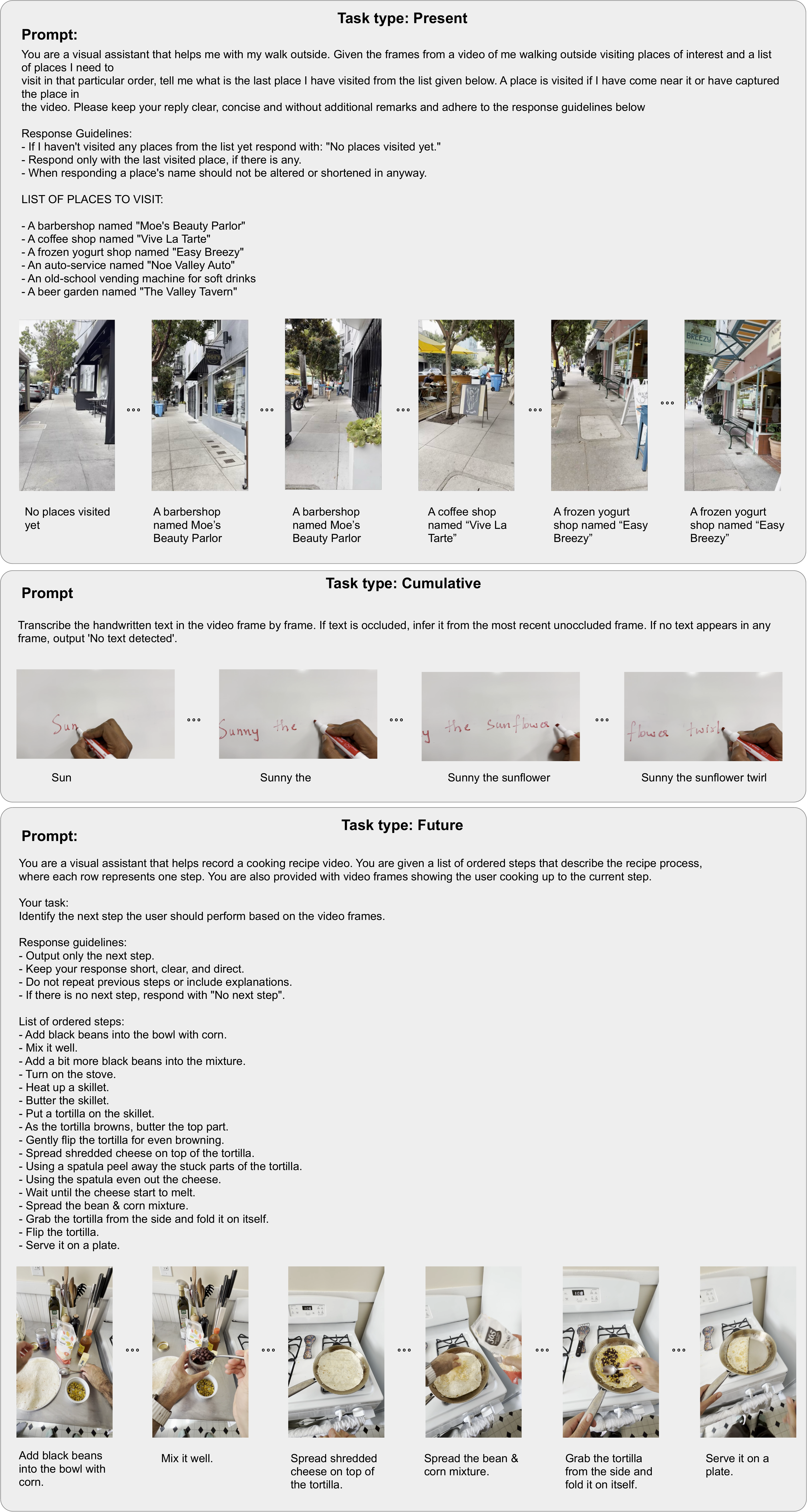}
    \vspace*{-5pt}
    \caption{\textbf{Examples of \ours{} task types with frame-level annotations and full prompts.} \ours{} involves three task types, \present{} tasks, which focus on currently occurring events; \cumulative{} tasks, which require the model to reason over past events; and \future{} tasks, which focus on predicting upcoming events based on ongoing visual cues.}
    \label{fig:benchmark_annotations_suppl}
\end{figure*}

We generate dense annotations using GPT-5, applying the sliding-window with uniform-tail (\SWUaccess{}) memory policy described in \cref{sec:video_to_streaming} with a buffer size of 42. Annotations are produced at 1 FPS, yielding free-form descriptions for every second of video. All auto-generated annotations are subsequently human-verified, with particular attention to maintaining consistency across adjacent frames, especially when the underlying scene remains unchanged. Additional examples with full prompts are provided in \cref{fig:benchmark_annotations_suppl}. Notably, our prompt design explicitly specifies the task granularity to minimize ambiguity and ensure that models respond at the appropriate level of detail.

\section{Detailed Comparison with Recent Streaming Benchmarks}
In \cref{tab:bench_comp}, we report dataset statistics in comparison to two recently introduced video benchmarks. Our benchmark is of comparable scale to prior benchmarks when evaluated under realistic streaming settings. Although it contains fewer source videos, it provides richer and denser annotations, with a median duration of 1.0s between annotations and an average of 11.8 unique events per video, compared to 61.8s and 2.5 unique events in RTV-Bench. This dense multi-timestamp QA annotation (MTQA) is critical for evaluating temporal fidelity and latency effects in streaming models. Moreover, streaming models must be evaluated on contiguous, overlapping video segments to reflect real-world usage; when evaluated in this mode, our benchmark incurs a comparable single-GPU evaluation cost to RTV-Bench, demonstrating similar effective evaluation scale. We note that these datasets are complementary to our streaming dataset.

\section{Robustness to category/task imbalance.}
Not every video in the benchmark supports all task types, as is also the case in prior benchmarks such as OVO-Bench and RTV-Bench. Some task types are intentionally omitted when they are not meaningful. For instance, the \future{} task is excluded in certain Text Understanding/OCR scenarios, where videos often involve egocentric reading and queries target text that is already visible or previously observed, rendering future prediction ill-defined. To evaluate the impact of this imbalance, we report results with Inverse-Category and Inverse-Task reweighting in \cref{tab:imbalance}. Model rankings and overall conclusions remain unchanged, suggesting that the benchmark is robust to variation in task composition.

\begin{table}[t]
\centering
\resizebox{1.0\columnwidth}{!}{%
    \setlength{\tabcolsep}{3pt}
    \begin{tabular}{l|ccc}
    \toprule
    Benchmark & OVO-Bench & RTV-Bench & \textbf{Ours} \\
    \midrule
    Number of video clips & \textbf{644} & 522 & 92 \\
    MTQA median. duration between annotations (sec.) & 4.0 & 61.8 & \textbf{1.0} \\
    Total Annotations  & 3207 & 4608 & \textbf{18410} \\
    Avg. unique events per video clip & 2 & 2.5 & \textbf{11.8} \\
    
    Total number of unique events & 3207 & \textbf{4608} & 2117 \\
    \midrule
    Single GPU eval. duration (mins.) & 127 & 395 & 345 \\
    \bottomrule
    \end{tabular}%
}
\caption{\small
\textbf{Benchmark Comparison} Single GPU eval duration is measured for Qwen2.5VL-7B-Instruct model.
}\label{tab:bench_comp}
    \vspace*{-5pt}
\end{table}

\begin{table}[t]
\centering
\resizebox{1.0\columnwidth}{!}{%
    \setlength{\tabcolsep}{5pt}
    \begin{tabular}{l|ccc|c|ccc|c}
    \toprule
    \multirow{2}{*}{Reweighting} & \multicolumn{4}{c}{Qwen3-VL-4B} & \multicolumn{4}{|c}{Qwen3-VL-32B} \\
    \cmidrule{2-9}
     & Present & Cumulative & Future & Overall & Present & Cumulative & Future & Overall \\
    \midrule
    
    \rowcolor{gray!15} Uniform (reported in main paper Tab. 2)  & 62.3 & 36.4 & 17.9 & 46.8 & 54.8 & 32.1 & 14.5 & 41.0 \\
    Inverse Category          & 64.6 & 38.5 & 26.4 & 48.4 & 58.1 & 37.3 & 19.9 & 43.5 \\
    Inverse Task              & 63.3 & 38.4 & 24.8 & 47.8 & 56.2 & 35.9 & 18.9 & 42.3 \\
    Inverse Category and Task & 62.3 & 43.4 & 17.3 & 45.8 & 54.7 & 41.1 & 13.7 & 41.1 \\
    
    \bottomrule
    \end{tabular}%
}
\caption{\small
Model accuracies for various reweighting schemes.
}\label{tab:imbalance}
    \vspace*{-10pt}
\end{table}

\section{Model-specific Optimizations}

Prior works like~\citep{xu2025streamingvlm, zhang2024flash} have demonstrated that model-specific optimizations like, KV-cache and embedding reuse strategies in streaming VLMs can substantially reduce inference latency by minimizing prefill time. To complement these approaches, we introduce an optimization technique inspired by speculative decoding methods in language modeling. This method extends speculative decoding to the streaming setting, enabling further reductions in latency without modifying or retraining the underlying model.

\subsection{Self-Speculative Decoding for Streaming}\label{sec:ssd} 
To reduce inference latency in the streaming setting, we introduce a variant of self-speculative decoding tailored for continuous video input. Unlike prior approaches that rely on lightweight draft models, we re-purpose the previous timestep’s generated tokens as the draft sequence and verify them against the current visual input. Because visual scenes in many streaming applications evolve gradually, the prior response is often still valid, allowing the model to approve the draft without generating new tokens. This verification step is highly efficient, as it can be executed in parallel and avoids full decoding. New tokens are produced only when verification fails, i.e., when the visual stream exhibits a meaningful change requiring an updated response. 

We evaluate this technique on a subset of text-recognition videos for cumulative task, with results summarized in \cref{tab:ssd}. Incorporating our variant of self-speculative decoding yields improvements across all metrics. The substantial reduction in inference latency leads to a 33.6\% gain in accuracy under asynchronous protocol, and consistency also increases because the model generates new outputs only when verification fails. This reduces unnecessary re-wordings across timesteps, thereby lowering the incidence of inconsistent responses.

\begin{figure}[t]
    \centering
    \includegraphics[width=1.0\columnwidth]{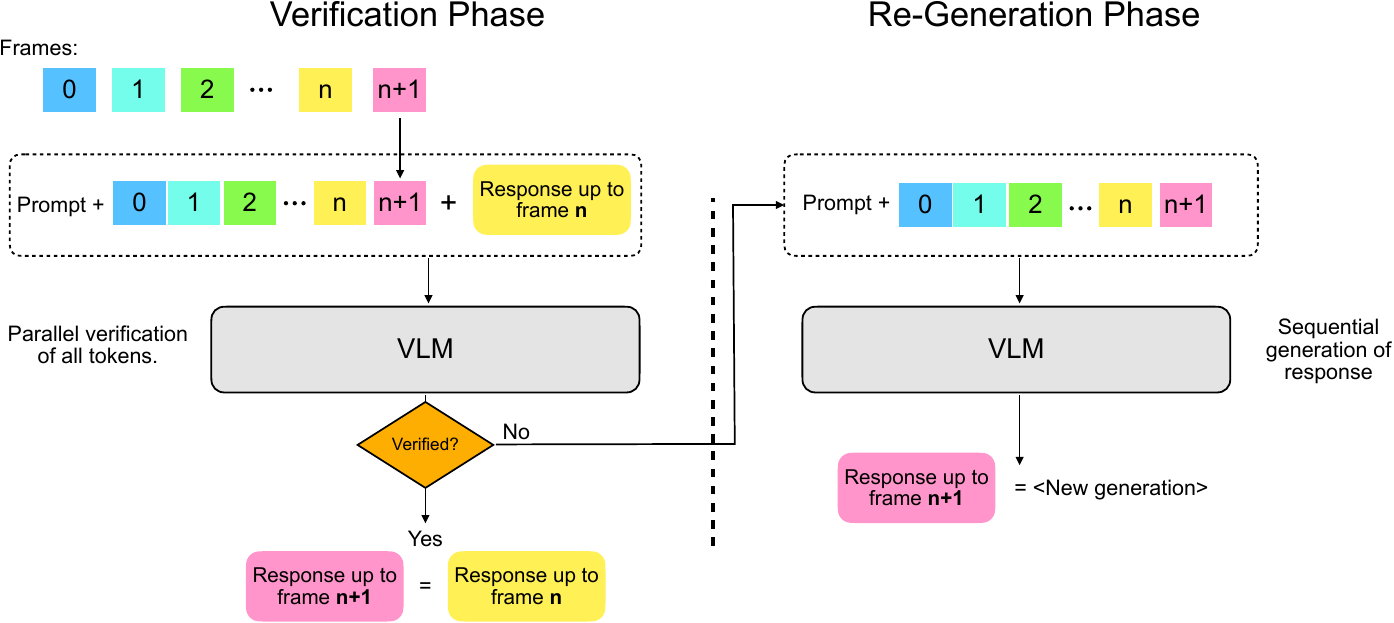}
    \caption{\textbf{Overview of self-speculative decoding for streaming VLMs}. The prior response serves as a draft to be verified for the next frame. 
}
    \label{fig:ssd_schema}
\end{figure}

\begin{table}[t]
\centering
\resizebox{0.85\columnwidth}{!}{%
\begin{tabular}{ccccc}
\toprule
 \multirow{2}{*}{\vspace{-0.2cm}\shortstack[c]{\textbf{Self-Speculative}\\\textbf{Decoding}}} & \multicolumn{2}{c}{\textbf{\cumulative{} Task} (OCR)} & \multirow{2}{*}{\vspace{-0.2cm}\shortstack[c]{\textbf{Mean Avg.}\\\textbf{Latency (s)}}} \\
\cmidrule(lr){2-3}
 & \textbf{Accuracy} & \textbf{Consistency}  &  \\
\midrule
\rowcolor{gray!15} \xmark & 21.5 & 93.0 & 5.8  \\
\cmark    & \textbf{55.1} & \textbf{96.2} & \textbf{1.5} \\
\bottomrule
\end{tabular}
}
\caption{\textbf{Self-Speculative Decoding for Streaming VLMs.}  We evaluate streaming-adapted Qwen3-VL-8B model using the asynchronous protocol and compare the performance of the model with and without self-speculative decoding optimization.}
\label{tab:ssd}
\vspace{-3mm}
\end{table}

\section{Accuracy evaluation using a Judge-LLM}\label{sec:eval_judge}
For accuracy evaluation, we use GPT-5 (medium reasoning)~\cite{openai2025gpt5} as a judge model to compare a model's output at each timestep with the reference ground-truth caption. We use a detailed judge prompt to clarify the scoring criteria for the different tasks in the benchmark. Specifically, the judge is designed to reduce variance across evaluation attempts (caused by different reasoning chains). The \texttt{<question>}, \texttt{<gt\_answer>}, and \texttt{<model\_response>} blocks at the end of prompt are replaced with the corresponding question, ground truth caption, and model response text.

\begin{figure*}[t]
\begin{tcolorbox}
You are the evaluator.\\
You will receive:\\
1) A user question\\
2) A model predicted response\\
3) A ground truth answer\\
Your task is to compare the model’s response with the ground truth and decide if they match meaningfully.\\

\#\#\#\#\#\#\#\#\#\#\# Instruction on how to evaluate \#\#\#\#\#\#\#\#\#\#\#\\
\\
\#\#\#\\
Your Output Format:\\
\\
Return only a JSON dictionary with the following keys:\\
\\
*  'pred': "yes" if the model response meaningfully matches the ground truth, "no" otherwise.\\
* 'score': an integer (not a string) between 0 and 3, based on how correct the model response is.\\
\\
Do not provide any explanation, notes, or text outside the JSON output.\\
Example output:\\
{'pred': 'yes', 'score': 2}\\
\\
\\
\#\#\#\\
Evaluation Rules:\\
\\
Binary Match (pred key)\\
\\
* Focus on meaningful equivalence between model response and ground truth.\\
* Accept synonyms, paraphrases, or reworded answers if meaning is preserved.\\
* Lists are acceptable if items are semantically aligned or paraphrased.\\
* All responses that qualify for Tier 3 (perfect match) are automatically labeled 'yes'.\\
* All responses that qualify for Tier 0 or 1 are automatically labeled 'no'.\\
\\
\\
\\
Rubric Score (score key)\\
\\
* Tier 0: No meaningful match.\\
* Tier 1: Some overlap, but major errors or missing key parts.\\
* Tier 2: Mostly correct with small mistakes or omissions.\\
* Tier 3: Perfect or near-perfect match for key elements in the question.\\
\\
\\
\#\#\#\\
\\
Specific Grading Guidelines\\
\\
* A response is considered a match if it includes the key elements asked in the question. For named entities, different spellings are acceptable.\\
\hspace*{1em}* Example: "Valley Tavern" is an acceptable match for ground truth "A beer garden named ‘The Valley Tavern’."\\
\hspace*{1em}* For general objects, synonyms are acceptable. Example: "shorts" is acceptable for "a pair of gray shorts" when the question only asks "what object."\\
\end{tcolorbox}
\end{figure*}

\begin{figure*}[t]
\begin{tcolorbox}
* If the model response is conceptually related to the ground truth, give partial credit.\\
\hspace*{1em}* Example: "backpack" instead of "handbag," "bottle of green tea" instead of "bottle of beer," or "cloth" instead of "t-shirt."\\
* When the question gives specific choices (e.g., "Crosswalk," "Sidewalk," or "Motorway"), the model response must exactly match one of the choices for a tier 3 score.\\
\hspace*{1em}* Minor spelling differences are fine.\\
\hspace*{1em}* A synonym not in the list (e.g., "Pavement" for "Sidewalk") gets tier 2.\\
\hspace*{1em}* Irrelevant responses get tier 0.\\
* For questions asking about an activity or cooking step (without choices):\\
\hspace*{1em}* Include all key elements : tier 3.\\
\hspace*{1em}* Miss some elements : tier 2.\\
\hspace*{1em}* Only vaguely capture a key element : tier 1.\\
\hspace*{1em}* Identify key elements based on the question.\\
\hspace*{1em}\hspace*{1em}* Example: If the question asks for the latest cooking step and ground truth is "Gather the ingredients and lay them out on the counter", then:\\
\hspace*{1em}\hspace*{1em}\hspace*{1em}* Response "Collect the ingredients" : tier 3.\\
\hspace*{1em}\hspace*{1em}\hspace*{1em}* Response "The person stops pointing at the ingredients and turns to speak to the camera" : tier 1.\\
\hspace*{1em}\hspace*{1em}* Example: If the question asks for the latest cooking step and ground truth is "Mix beans and olive oil well.", then:\\
\hspace*{1em}\hspace*{1em}\hspace*{1em}* Response "Add corn into the mixture." : tier 0.\\
\hspace*{1em}\hspace*{1em}\hspace*{1em}* Response "Add black beans into the mixture." : tier 1.\\
\hspace*{1em}\hspace*{1em}\hspace*{1em}* Response "Combine well." : tier 2.\\
\hspace*{1em}\hspace*{1em}\hspace*{1em}* Response "Blend black beans and oil thoroughly." : tier 3.\\
* When the question asks about a step from a given list of steps (e.g., for cooking or for computer tasks):\\
\hspace*{1em}* Model response exactly match ground truth or is its paraphrase capturing key element: tier 3.\\
\hspace*{1em}* Model response match ground truth but miss one important element: tier 2.\\
\hspace*{1em}* Irrelevant responses get tier 0.\\
\hspace*{1em}* For these cases do not assign score 1.\\
\hspace*{1em}* Example: Ground truth "Click the Safari icon in the Dock to open Safari."\\
\hspace*{1em}\hspace*{1em}* Response "Open Safari." : tier 3.\\
\hspace*{1em}\hspace*{1em}* Response "Click on the icon in the Dock." : tier 2.\\
\hspace*{1em}\hspace*{1em}* Response "Click on the button." : tier 0.\\
* If the model response misses an important part of the ground truth, give partial credit.\\
\hspace*{1em}* Example: Ground truth "bowl of rice."\\
\hspace*{1em}\hspace*{1em}* Response "bowl." : tier 1.\\
\hspace*{1em}\hspace*{1em}* Response "rice." : tier 2.\\
* Empty strings, "None," or "NA" (and similar responses) are considered a match.\\
* When the ground truth is a list and order is not important, grade based on the intersection-over-union (IOU) of items:\\
\hspace*{1em}* All items match : tier 3.\\
\hspace*{1em}* Most items match ($1>$IOU$>0.8$) : tier 2.\\
\hspace*{1em}* Few items match ($0<$IOU$<0.8$) : tier 1.\\
\hspace*{1em}* No match : tier 0.\\
\hspace*{1em}* Lists can be comma-separated, space-separated, or multiline.\\
\hspace*{1em}* Example: Ground truth "Bowl of rice, beer bottle." and response "bowl, carrot" : tier 1.\\
\end{tcolorbox}
\end{figure*}
\begin{figure*}[t]
\begin{tcolorbox}
* When the ground truth is a list and order is important (e.g., question asks for chronological order):\\
\hspace*{1em}* All items match with the same order as ground truth : tier 3.\\
\hspace*{1em}* All items match with the ground truth but order is different : tier 2.\\
\hspace*{1em}* Some items match with the ground truth : tier 1.\\
\hspace*{1em}* No match : tier 0.\\
* For transcription questions:\\
\hspace*{1em}* Tier 3: Exact match (minor character-level differences acceptable).\\
\hspace*{1em}* Tier 2: Mostly correct; less than 20\% of words missing or changed.\\
\hspace*{1em}* Tier 1: Partially correct; more than 20\% of words missing, changed, or added.\\
\hspace*{1em}* Tier 0: Not following the ground truth.\\
* For counting questions:\\
\hspace*{1em}* Exact number match : tier 3 (numeric or word form both acceptable).\\
\hspace*{1em}* Format matches but count is wrong : tier 1.\\
\hspace*{1em}* Format also incorrect : tier 0.\\
\\
\#\#\#\#\#\#\#\#\#\#\# Data for evaluation \#\#\#\#\#\#\#\#\#\#\#\\
\\
Please evaluate the following video-based question-answer pair:\\
\\
Question:\\
$<$question$>$\\
\\
Ground truth answer:\\
$<$gt\_answer$>$\\
\\
Model predicted response:\\
$<$model\_response$>$\\
\end{tcolorbox}
\caption{Judge prompt used to evaluate accuracy of model response.}
\end{figure*}

\section{Compute Cost Analysis}

We evaluate models smaller than 8B on a single video using a single GPU, while sharding 8B models on 2 GPUs and 32/38B models on 4 GPUs. We distribute the evaluation on different videos across a cluster of GPUs. Evaluating Qwen3-VL-2B-Instruct on all videos distributed over 8 GPUs takes 1 hour (8 GPU hours), while evaluating Qwen3-VL-8B-Instruct takes 1.5 hours (12 GPU hours). We utilize H100 GPUs (AWS instance type p5.48xlarge) for all evaluations.

\section{End-to-End Latency per Task}
We report average end-to-end latency in \cref{tab:latency}. Under the asynchronous protocol, accuracy reflects the trade-off between computational delay and temporal fidelity, as delayed responses are evaluated against later frames (\cref{sec:async_protocol}). GPT-5 is accessed via web API, and its associated network latency degrades performance under true streaming conditions compared to locally deployed models such as Qwen3VL and InternVL3.

\begin{table}[h]
\centering
\resizebox{1.0\columnwidth}{!}{%
    \setlength{\tabcolsep}{2pt}
    \begin{tabular}{l|ccc|cc|c}
    \toprule
     & Qwen3VL-4B & Qwen3VL-8B & Qwen3VL-32B & FlashVStream (7B) & Dispider (7B) & GPT-5$^{\dagger}$\\
    \midrule
    \rowcolor{gray!15} Present & 1.3 & 1.6 & 2.4 & 0.8 & 1.7 & 42.4 \\
    Cumulative & 2.1 & 2.2 & 4.1 & 1.4 & 1.2 & 63.7 \\
    \rowcolor{gray!15} Future & 2.1 & 2.7 & 3.5 & 0.3 & 3.3 & 65.8 \\
    \bottomrule
    \end{tabular}%
}\vspace*{-5pt}
\caption{\small
Average end-to-end latency in seconds per task-type. Video and streaming models were evaluated on NVIDIA H100. $\dagger$ For GPT-5 its the measure of its API latency.
}\label{tab:latency}
\end{table}

\begin{table}{
\resizebox{1.0\columnwidth}{!}{
\begin{NiceTabular}{c|cccc|c}
    \toprule[1.5pt]
    \multirow{2}{*}{\textbf{Model}} 
        & \multicolumn{4}{c}{\textbf{Async. Accuracy} $\uparrow$}
        & \multirow{2}{*}{\shortstack{\textbf{Async.}\\\textbf{Consistency} $\uparrow$}} \\
    \cmidrule(lr){2-5}
      & \textbf{\present{}} & \textbf{\cumulative{}} & \textbf{\future{}} & \textbf{Overall}
      & \\
    \midrule[1.25pt]
    Qwen3-VL-2B~\citep{qwen3vl}                     & 54.0 & 29.9 & 9.6 & 39.0 & 96.5 \\
    \rowcolor{gray!15} Qwen3-VL-4B~\citep{qwen3vl}  & \textbf{61.6} & 39.3 & 18.4 & \textbf{47.4} & 95.0 \\
    Qwen3-VL-8B~\citep{qwen3vl}                     & 57.8 & \textbf{41.6} & \textbf{22.0} & 46.8 & 95.8 \\
    \rowcolor{gray!15} Qwen3-VL-32B~\citep{qwen3vl} & 55.6 & 36.1 & 15.6 & 42.8 & \textbf{96.7} \\
    \bottomrule[1.5pt] 
\end{NiceTabular}
}
\caption{Async protocol performance for Qwen3-VL family of models evaluated with \SWUaccess{} memory policy.}\label{tab:supp_best_results}
}\end{table}

\begin{table}[t]
\centering
\resizebox{0.82\columnwidth}{!}{%
\begin{tabular}{ccccc}
\toprule
 \multirow{2}{*}{\vspace{-0.2cm}\shortstack[c]{\textbf{Camera}\\\textbf{Buffer Size}}} & \multicolumn{3}{c}{\textbf{Tasks}} & \multirow{2}{*}{\textbf{Overall}}  \\
\cmidrule(lr){2-4}
 & \textbf{\present{}} & \textbf{\cumulative{}} & \textbf{\future{}} &  \\

\midrule
\multicolumn{5}{c}{\textbf{Qwen3-VL-2B}} \\
\midrule
 \rowcolor{gray!15}  1   & 53.8 & 28.4 & 9.6  & 38.5 \\
                     8   & 54.2 & 30.4 & 9.8  & 39.3 \\
\rowcolor{gray!15}   16  & \textbf{54.3} & \textbf{31.2} & 10.1 & \textbf{39.7} \\
                     600 & 53.6 & 30.5 & \textbf{10.5} & 39.1 \\

\midrule
\multicolumn{5}{c}{\textbf{Qwen3-VL-8B}} \\
\midrule
 \rowcolor{gray!15}  1   & 58.4 & 37.5 & 21.5 & 45.7 \\
                     8   & 58.0 & 39.8 & 22.5 & 46.4 \\
 \rowcolor{gray!15}  16  & 57.8 & 41.1 & 21.9 & 46.6 \\
                     600 & \textbf{62.4} & \textbf{52.6} & \textbf{35.1} & \textbf{54.8} \\

\bottomrule
\end{tabular}
}
\vspace*{-3pt}
\caption{\textbf{Effect of camera buffer size.} We evaluate Qwen3-VL 2B and 8B sizes using the asynchronous protocol. All runs use memory buffer size 64 and \SWUaccess{} policy. We highlight the best numbers in each column.}
\label{tab:abl_camera_buffer}
\end{table}

\section{Results with \SWUaccess{} Memory Policy}
In \cref{tab:main_results}, all streaming-adapted video VLMs evaluated under the asynchronous protocol use the sliding-window (\SWaccess{}) memory policy with a buffer size of 64. Here, we report results for streaming-adapted Qwen3-VL models using the sliding-window with uniform tail (\SWUaccess{}) policy, which provides a better balance between recency and long-range temporal coverage. As shown in \cref{tab:supp_best_results}, SW+U consistently improves performance, particularly on cumulative tasks; for instance, Qwen3-VL-4B gains 5.2\% in cumulative accuracy.

\section{Effect of Camera Buffer Size}
In \cref{tab:main_results}, all models evaluated under the asynchronous protocol use a camera buffer of size 600, which is large enough to exceed the maximum number of frames in any video, ensuring that no frames are dropped and accuracy is influenced solely by model latency. However, in a realistic setup, systems cannot assume an unbounded camera buffer. We therefore ablate performance under varying camera buffer sizes, which induces frame drops when the model is too slow to keep pace with the incoming stream. This analysis highlights how constrained buffering alters a model’s effective temporal context under the same memory policy. Our implementation exposes this parameter, allowing practitioners to adjust it to match their deployment constraints. In \cref{tab:abl_camera_buffer}, we report the performance of streaming-adapted Qwen3-VL models for various camera buffer sizes. For models such as Qwen3-VL-8B, reducing the camera buffer to 16 leads to a substantial degradation, with overall accuracy dropping by nearly 15\%.

\end{document}